\documentclass[final,5p, times]{elsarticle}

\usepackage{graphicx}
\usepackage{amssymb}
\usepackage{multirow}
\usepackage{hyperref}

\usepackage{lineno}

\usepackage{multirow}
\usepackage{amssymb}
\usepackage{subfig}
\usepackage{float}
\usepackage{lscape}
\usepackage{multirow}
\usepackage{amsmath}
\usepackage{array}
\usepackage{textcomp}
\usepackage{multirow}
\newcolumntype{L}[1]{>{\raggedright\let\newline\\\arraybackslash\hspace{0pt}}m{#1}}
\newcolumntype{C}[1]{>{\centering\let\newline\\\arraybackslash\hspace{0pt}}m{#1}}
\newcolumntype{R}[1]{>{\raggedleft\let\newline\\\arraybackslash\hspace{0pt}}m{#1}}
\usepackage{tabularx} 
\usepackage{lscape}
\usepackage{longtable}
\usepackage[mathscr]{euscript}
\usepackage[]{dcolumn}
\usepackage[noabbrev,capitalise]{cleveref}
\usepackage{soul}

\usepackage{enumitem}

\usepackage{xcolor}
\hypersetup{
    colorlinks,
    linkcolor={blue},
    citecolor={blue},
    urlcolor={blue}
}

\modulolinenumbers[1]

\biboptions{authoryear}

\journal{-}

\def \resuneta {\texttt{ResUNet-a} }
\def \dsix {\texttt{ResUNet-a D6} }
\def \dseven {\texttt{ResUNet-a D7} }

\newcommand{\ie}{\textit{i.e.}, }
\newcommand{\eg}{\textit{e.g.}, }

\usepackage{setspace}
\begin{document}
\begin{frontmatter}


% where the fields end
\title{Deep learning on edge: extracting field boundaries from satellite images with a  convolutional neural network}
% Finding field boundaries in satellite images with a deep conv
%\title{A deep convolutional neural network that extracts field boundaries from satellite images and generalises across space, time, resolutions and sensors}
%\title{Extracting field boundaries from satellite imagery with a cutting-edge, deep convolutional neural network}


\author[label1]{Fran\c{c}ois Waldner \corref{cor1}}
\author[label2]{Foivos I. Diakogiannis}

\address[label1]{CSIRO Agriculture \& Food, 306 Carmody Road, St Lucia, Queensland, Australia}
\address[label2]{CSIRO Data61, Analytics, 147 Underwood Avenue, Floreat, Western Australia, Australia}

\cortext[cor1]{Corresponding author; E-mail: franz.waldner@csiro.au}

\begin{abstract}
Applications of digital agricultural services often require either farmers or their advisers to provide digital records of their field boundaries. Automatic extraction of field boundaries from satellite imagery would reduce the reliance on manual input of these records which is time consuming, and would underpin the provision of remote products and services. The lack of current field boundary data sets seems to indicate low uptake of existing methods, presumably because of expensive image preprocessing requirements and local, often arbitrary, tuning. In this paper, we sought to facilitate field boundary extraction from satellite images by proposing a data-driven, robust, general method and addressed this problem as a multi-task semantic segmentation problem. We used \texttt{ResUNet-a}, a deep convolutional neural network with a fully connected UNet backbone that features dilated convolutions  and conditioned inference to identify:  1) the extent of  fields; 2) the field boundaries; and 3) the distance to the closest boundary. By asking the algorithm to reconstruct three correlated outputs, model performance and its generalisation can greatly improve. Next, instance segmentation of individual fields can be achieved by post-processing the three model outputs, \eg via thresholding or watershed segmentation. Using a single monthly composite image from Sentinel-2 as input, the model was highly accurate in mapping field extent, field boundaries, and, consequently, individual fields. Replacing the monthly composite with a single-date image close to the compositing period  marginally decreased accuracy. We then showed in a series of experiments that,  without recalibration, the same model generalised well across resolutions, sensors, space and time.  Building consensus by averaging model predictions from at least four images acquired across the season is the key to coping with the temporal variations of accuracy. The convolutional network is capable of learning complex  hierarchical contextual features from the image to accurately detect  field boundaries and discard irrelevant boundaries, thereby outperforming conventional edge filters. By minimising over-fitting and image preprocessing requirements, and by replacing local arbitrary decisions by data-driven ones, our approach is expected to facilitate the extraction of individual crop fields at scale. 
\end{abstract}

\begin{keyword}
Agriculture \sep Field boundaries \sep Sentinel-2 \sep  Semantic segmentation \sep Instance Segmentation \sep Multitasking \sep Computer vision \sep Generalisation


\end{keyword}

\end{frontmatter}


%\linenumbers

\section{Introduction}

% https://www.researchgate.net/publication/319618257_A_Novel_Road_Segmentation_Technique_from_Orthophotos_Using_Deep_Convolutional_Autoencoders

Many of the promises of digital agriculture centre on assisting farmers to monitor their fields throughout the growing season. Having precise field boundaries has become a prerequisite for field-level assessment and often, when farmers are being signed up by service providers, they are asked for precise digital records of their boundaries. Unfortunately, this process remains largely manual and time-consuming, which creates disincentives.  There are also increasing applications whereby remote monitoring of crops using earth observation is used for estimating areas of crop planted yield forecasts as well as for monitoring food security~\citep{de2004efficiency, blaes2005efficiency, matton2015automated}. These applications would  greatly benefit from repeated extraction of field boundaries across large areas. Automating this process not only facilitates bringing farmers on board, and hence fostering wider adoption of digital agricultural services, but also allows improved products and services to be provided using remote sensing.\\

%Visual image analysis coupled with on-screen digitisation remains probably the most straightforward and trusted approach to obtain field boundaries. But manual digitisation does not scale, is time-consuming and error-prone.

%Beyond digital agricultural services, the spatial distribution of fields inform about agriculture mechanisation~\citep{kuemmerle2013challenges},  human development, species richness~\citep{geiger2010persistent}, resource allocation and economic planning~\citep{carfagna2005using,rudel2009agricultural,johnson20132010}. Field boundaries also support a range of remote-sensing applications such as selecting the map or imagery with the most appropriate spatial resolution~\citep{waldner2015mapping, waldner2017can}, classifying crops~\citep{de2004efficiency}, and reducing speckle in radar images or crop type classifications~\citep{blaes2005efficiency, matton2015automated}.  

% https://www.spiedigitallibrary.org/journals/Journal-of-Applied-Remote-Sensing/volume-3/issue-1/033528/Potential-of-IRS-P-6-LISS-IV-for-agriculture-field/10.1117/1.3133306.short?SSO=1
% http://www.isprs.org/proceedings/XXXVIII/4-C1/Sessions/Session10/6671_Smith_Proc_pap.pdf
% https://www.tandfonline.com/doi/abs/10.1080/07038992.1988.10855126

Several approaches have been devised to extract field boundaries from satellite imagery, which provides regular and global coverage of cropping areas at high resolution. These approaches generally fall  into three categories: edge-based methods~\citep{mueller2004edge, shrivakshan2012comparison, turker2013field, yan2014automated, graesser2017detection}; region-based methods~\citep{evans2002segmenting, mueller2004edge, salman2006image, garcia2017machine}; and hybrid methods that seek to address shortcomings of one category with components from the other~\citep{rydberg2001integrated}. Edge-based methods rely on filters  to identify discontinuities in images where pixel values change rapidly. Each filter defines a specific kernel (the Scharr, Sobel, and Canny operators are common examples) which is then convolved with the input image to emphasise edges. A number of issues arise when working with edge operators: their sensitivity to high-frequency noise often creates false edges; and their parameterisation is arbitrary, relevant to specific contexts, and a single parameterisation may lead to incomplete boundary extraction. Post-processing and locally-adapted thresholds may  address these issues leading to better-defined, closed boundaries. Region-based algorithms, such as the watershed segmentation \citep{soille1990automated}, group neighbouring pixels into objects based on some homogeneity criterion. Finding the optimal segmentation parameters for region-based algorithms remains a trial and error process that likely yields sub-optimal results. For instance, with poor parameterisation, objects may stop growing before reaching the actual boundaries~\citep{chen2015image}, creating sliver polygons and shifting the extracted boundaries inwards. Region-based methods also tend to over-segment fields with high internal variability and under-segment small adjacent fields~\citep{belgiu2018sentinel}. Some of these adverse effects might be mitigated by purposefully over-segmenting images and deciding whether adjacent objects should be merged with machine learning~\citep[see][for instance]{garcia2017machine}. Despite the availability of edge-based and region-based methods, there seems to be a low uptake of these methods by the user community, suggesting a lack of fitness for purpose. For instance, the only global map of field size was obtained from crowdsourced, manually-digitised polygons~\citep{lesiv2019estimating}.\\

% depending on the task, some produce finer edges than others~\citep{watkins2019comparison}. 

%\citet{rydberg2001integrated} presented a multispectral segmentation method for automated delineation of agricultural field boundaries in remotely sensed images. Edge information from a gradient edge detector is integrated with a segmentation algorithm. 

This low uptake can  be explained by  the expense of image preprocessing, local, often arbitrary, tuning which does not generalise to other locations, and requires ancillary data, such as cropland or crop type maps~\citep[\eg][]{yan2014automated, graesser2017detection}. Deriving multitemporal image features seem unnecessary because, in many cases, humans  can to draw field boundaries from well-targeted single-date images. Besides, persistent cloud coverage  makes it difficult to generate consistent time series in areas such as the tropics.  While multitemporal information is likely  to improve accuracy especially in those highly-dynamic systems, we think that new methods with lower preprocessing requirements and data-driven parameter tuning are needed to facilitate large-scale  extraction of field boundaries and improve their uptake. \\

Deep neural networks open up new avenues for field boundary extraction because they do not require  hand-crafted features and their architectures are highly adaptive to new problems~\citep{agravat2018deep}. Deep convolutional neural networks, a group of deep neural networks which use convolution operations, are increasingly used in image analysis because they can exploit  hierarchical (local to global) features in images. While filters are hand-engineered in edge-based methods, convolutional neural networks can learn these filters. Initially devised for natural images, these networks have been revisited and adapted to tackle semantic segmentation problems (the process of linking each pixel in an image to a class label) in remote sensing, such as road extraction~\citep{cheng2017automatic}, cloud detection~\citep{chai2019cloud}, crop identification~\citep{ji20183d}, river and water body extraction~\citep{chen2018extraction, isikdogan2018learning}, and urban mapping~\citep{diakogiannis2019resunet}. As such, convolutional neural networks seem particularly well-suited to extract field boundaries at scale, but this has yet to be empirically proven. \\ 

% Deep neural networks techniques [9], in particular, the deep convolutional neural networks (DCNN) [10], have dramatically improved the state-of-the-art in object categorization and object detection. The DCNN has also been widely used on biomedical dataset, such as for skin lesion analysis [11, 12].

%However we cannot know how many different instances of the same class exist. Instance segmentation is semantic segmentation with the additional information that we can distinguish between different objects of the same type.  Therefore, instance segmentation will automatically extract: our methodology extracts: a) the class label per pixel, b) how many different objects (i.e. paddocks) exist in an image.}

Our overarching aim is to develop and evaluate a method to routinely extract field boundaries at scale by both replacing context-specific arbitrary decisions and minimising the image preprocessing workload.  We formulated this task as a multi-task semantic segmentation problem for a fully convolutional neural network where each pixel is simultaneously annotated with three labels: 1) the probability of belonging to a field (extent mask); 2) the probability of belonging to a boundary (boundary mask); and 3) the distance to the closest boundary (distance mask). We used  \texttt{ResUNet-a} \citep[a deep convolutional neural network with a fully connected UNet architecture which features dilated convolution and conditioned inference;][]{diakogiannis2019resunet} to extract field boundaries\footnote{See \url{https://github.com/feevos/resuneta} for a python implementation of the \texttt{ResUNet-a} model.}. As the neural network might produce discontinuous boundaries, we also introduce two post-processing methods that leverage its outputs to generate closed boundaries and retrieve individual fields. In other words, the post-processing methods exploit the semantic segmentation outputs to achieve instance segmentation.\\

This paper intends to test the performance and limitations of the proposed approach and to lay the blueprint for national to global field boundary extraction using deep learning. As this paper will show, our method extracted field boundaries with high thematic and geometric accuracy when using a monthly composite image of Sentinel-2 across our main site in South Africa.  We then conducted a series of experiments which demonstrate that over-fitting was minimised, allowing our convolutional neural network to be applied across a range of conditions without recalibration. Specifically, it generalised well 1) to a single-date Sentinel-2 image close to the compositing period, 2) to a Landsat image, and 3) to other locations (secondary sites in Argentina, Australia, Canada, Russia, Ukraine) and acquisition dates. By learning spectral and contextual information, our deep convolutional neural network discards edges that are not part of field boundaries and emphasises those that are,  providing a clear advantage over conventional  methods. \\

This paper is organised as follows. In section~\ref{sec:method}, we describe our method to extract field boundaries (the neural network, the data augmentation procedure and the strategy for training and inference) and to post-process its outputs to retrieve individual fields (instance segmentation). Section~\ref{sec:data} presents the data collected for the main  and the secondary sites and section~\ref{sec:experiment} details the experimental design. Results are presented in Section~\ref{sec:results} and discussed in Section~\ref{sec:discussion}, where we propose evidence-based recommendations for large-scale field boundary extraction with deep learning.

% https://books.google.be/books?hl=fr&lr=&id=bRHLBAAAQBAJ&oi=fnd&pg=PA127&dq=%22field+boundary%22+segmentation+satellite&ots=0smSX7Stxv&sig=jeEmYkQMl0ChJ_ZIYrehdieMLJA#v=onepage&q&f=false

% The recent publicity surrounding homelessness has given (a) fresh impetus to the cause.

\section{Extracting field boundaries with multi-task semantic segmentation}
\label{sec:method}

Conceptually, extracting field boundaries consists of labelling each pixel of a multi-spectral image with one of two classes: ``boundary'' or ``not boundary''. While  predicting only the classes of interest (single-tasking) can achieve acceptable performance~\citep[see][ for instance]{persello2019delineation}, it ignores training signals of related learning tasks that might help  improve the accuracy of the initial task~\citep{ruder2017overview}. By sharing representations between related tasks \ie multi-task learning, a model can learn to generalise better on the initial task. Instead of predicting each related task simultaneously and independently, these  can also be combined in the last layer of the architecture to further constrain inference of the initial task (conditioned multitasking), which was found to reduce the variance,  to stabilise the gradient updates and thus to improve model performance~\citep{diakogiannis2019resunet}. \\

Therefore, we formulate the extraction of field boundaries as a semantic segmentation problem where the goal is to predict multiple class labels. We trained a convolutional neural network called \texttt{ResUNet-a} to perform the four correlated tasks: to map the extent of fields, to identify field boundaries, to estimate the distance to the nearest boundary and reconstruct input images (Fig.~\ref{subfig:architecture}; Section~\ref{sec:resuneta}).  We refer to \citet{brodrick2019uncovering} for a thorough introduction to convolutional neural networks and to \citet{diakogiannis2019resunet} for more details about the \texttt{ResUNet-a} framework.  \\

Our overarching goal is  to generate individual fields (instance segmentation) to enable per-field analytics, which requires that boundaries are closed contours. However, it is likely that a convolutional neural network provides open contours, especially if it is trained to predict only boundaries. Compared to single-tasking, multitasking is  likely to provide improved boundaries because it learns to predict multiple highly correlated outputs (boundary, distance, extent) and it provides more information that can help remedy the problem of open contours.  We introduce two data-driven post-processing methods to generate closed field boundaries and, therefore, to extract individual fields (Section~\ref{sec:postprocessing}).  One is based on thresholding and the other on watershed segmentation; both leverage  multiple outputs of \texttt{ResUNet-a}.

%\Foivos{Here (write it better), instead of requesting that the CNN produces only boundary segmentation maps, we require our algorithm to provide three highly correlated outputs, namely boundary, distance, extent. These outputs can help remedy the problem of open contours. Then ...}

\subsection{Boundary detection with a deep convolutional neural network}
\label{sec:resuneta}

\subsubsection{Model architecture}

\begin{figure*}[h!!!]
  \centering
  \subfloat[]{\label{subfig:causal}\includegraphics[width=0.5\linewidth]{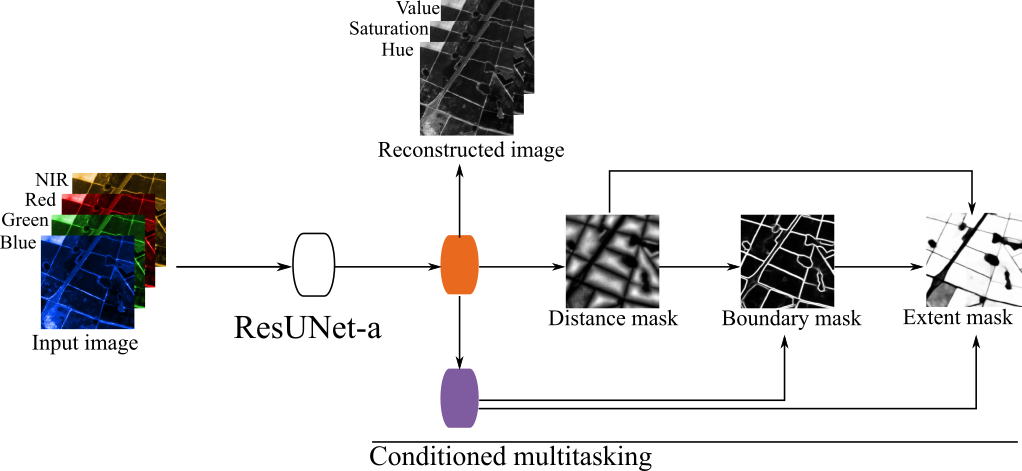}} \\
  \subfloat[]{\label{subfig:architecture}\includegraphics[width=0.8\linewidth]{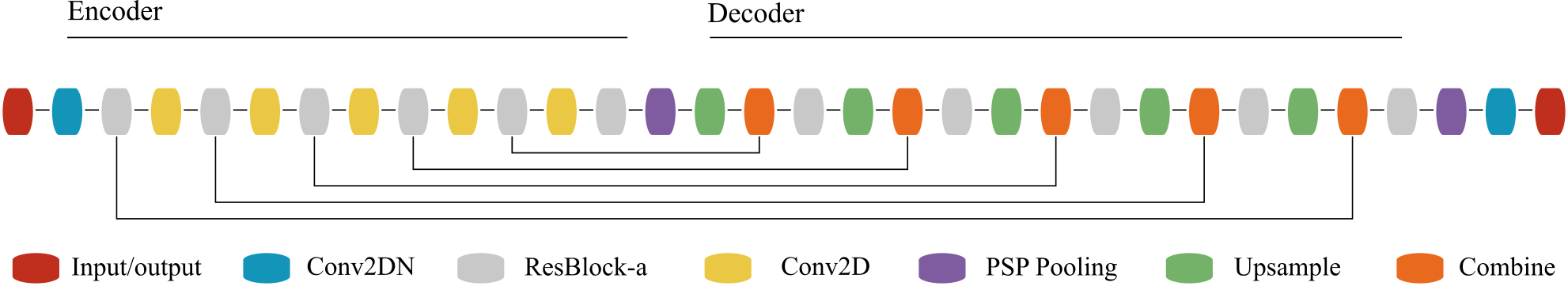}}
%  \subfloat[]{\label{subfig:double_watershed}\includegraphics[width=0.18\linewidth]{figures/watershed_boundary.png}}\hspace{0.5cm}
  \caption{\label{fig:architectures} Field boundary extraction formulated as multiple semantic segmentation tasks. (a) Overview of the \texttt{ResUNet-a}'s graph for multitasked inference. The distance mask is first generated  from the feature map, then is combined to the feature map to predict the boundary mask. Both masks are used to predict the extent mask. An independent branch reconstructs the input image. (b) Architecture of \texttt{ResUNet-a D6}}
\end{figure*}

The \texttt{ResUNet-a} model is a deep convolutional neural network, which has been previously shown to outperform state-of-the-art  architectures for semantic segmentation~\citep{diakogiannis2019resunet}. \texttt{ResUNet-a}  features the following components (Fig.~\ref{subfig:architecture}): 1) a UNet  architecture, which consists of a contracting path that captures context and a symmetric expanding path that localise objects precisely~\citep{ronneberger2015u}; 2) residual blocks, which helps alleviate the problem of vanishing and exploding gradients~\citep{he2016identity}; 3) atrous convolutions (with a range of dilation rates) that increase the receptive field~\citep{chen2017rethinking, chen2018deeplab}; 4)  pyramid scene parsing pooling to include contextual information~\citep{zhao2017pyramid}; and 5) conditioned multitasking. \texttt{ResUNet-a} produces four output layers: the extent mask, the boundaries, the distance mask \citep{Borgefors:1986:DTD:17140.17147}, and full reconstruction of the input image in the Hue-Saturation-Value colour space (Fig.~\ref{subfig:causal}). The colour space transformation provides additional information for the correlation between colour variations and object extent.\\

The UNet backbone architecture, also known as encoder-decoder, consists of two parts: the contraction part or encoder,  and the symmetric expanding path or decoder. The encoder compresses the information content of an arbitrarily high-dimensional image.  The  decoder gradually upscales the encoded features back to the original resolution and precisely localises the classes  of interest. The building blocks of  the encoder and decoder in the \texttt{ResUNet-a} architecture consist of residual units with  multiple parallel branches of atrous convolutions, each with a different dilation rate.\\

We implemented two basic architectures that differ in  depth (number of layers in the encoder-decoder). \dsix has six residual building blocks  in the encoder followed by a PSPPooling layer (Fig.~\ref{subfig:architecture}), \dseven has seven building blocks  in the encoder.\\

The different types of layers of \texttt{ResUNet-a} have the following functions:
\begin{description}
 \item[Input layer] stores the input image, which is a 256$\times$256 raster with four spectral bands (blue, green, red, near-infrared).
 
  \item[Conv2D layer] A standard 2D convolution layer (kernel size = 3, padding=1).
  
 \item[Conv2DN layer] This is a 2D convolution layer (kernel size = 3 and padding = 1) followed by a batch normalisation layer. By normalising the output of a previous activation layer by subtracting the batch mean and dividing by the batch standard deviation, batch normalisation is an efficient technique to combat the internal covariate shift problem and thus improves the speed, performance, and stability of artificial neural networks~\citep{ioffe2015batch}.
 
 \item[ResBlock-a layer] This layer follows the philosophy of the residual units  \citep{DBLP:journals/corr/HeZR016}, \ie units that skip connections between layers to improve the flow of information between the first and the last layers. Instead of having a single residual branch that consists of two successive convolution layers, there are up to four parallel branches with increasingly larger  dilation rates so that the input is simultaneously processed at multiple fields of view. The size of the input feature map dictates the dilation rates.
 
 \item[PSPPooling layer] The pyramid scene parsing pooling layer  emphasises contextual  information by applying maximum pooling at four different scales~\citep{zhao2017pyramid}. The first scale is a global max pooling. At the second scale, the feature map is divided into four equal areas and max pooling is performed in each of these areas, and so on for the next two scales. As a result, this layer encapsulates information about the dominant contextual features across scales.
 
 \item[Upsample layer] This layer consists of an initial upsampling of the feature map (interpolation) that is then followed by a  Conv2DN layer. The size of the feature map is doubled and the number of filters in the feature map is halved. 
 
 \item[Combine layer] This layer receives two inputs with the same number of filters in each feature map. It concatenates them and produces an output at the same scale, with the same number of filters as each of the input feature maps.   
 
 \item[Output layer] This is a multitasked layer with conditioned inference that produces a total of four output layers (Fig.~\ref{subfig:causal}): the extent mask, the boundary mask, the distance mask, and the reconstructed image of the input image  in the Hue-Saturation-Value space.  A network graph is constructed so as to take advantage of the inference results for the previous layers.  The process starts by combining the first and the last layers of the UNet. The distance mask is the first output to be generated without the use of PSPPooling. Then, to produce the boundary mask, the distance transform is conflated with the last UNet layer  using PSPPooling. The distance mask and the boundary mask are combined with the output features to generate the extent mask. A parallel fourth layer reconstructs the initial input image which also helps reduce the model variance~\citep{diakogiannis2019resunet}.

 %\Foivos{This is a conditioned, multitasking layer (Fig.~\ref{subfig:causal}). It produces a total of four output layers. These are:  This process takes place as follows: first the first and last layers of the UNet backbone are combined together. Then follows the color reconstruction, and the distance transform without the use of PSPPooling. The distance transform is being used in combination with the last extracted features (that are subject to PSPPooling) in order to produce the boundary logits. And, finally, the distance transform and the boundary layer are recombined with the last feature map and the extent mask is produced. }
\end{description}

\subsubsection{Data Augmentation}

\begin{figure}[h!!!]
\centering
\includegraphics[width=0.9\linewidth]{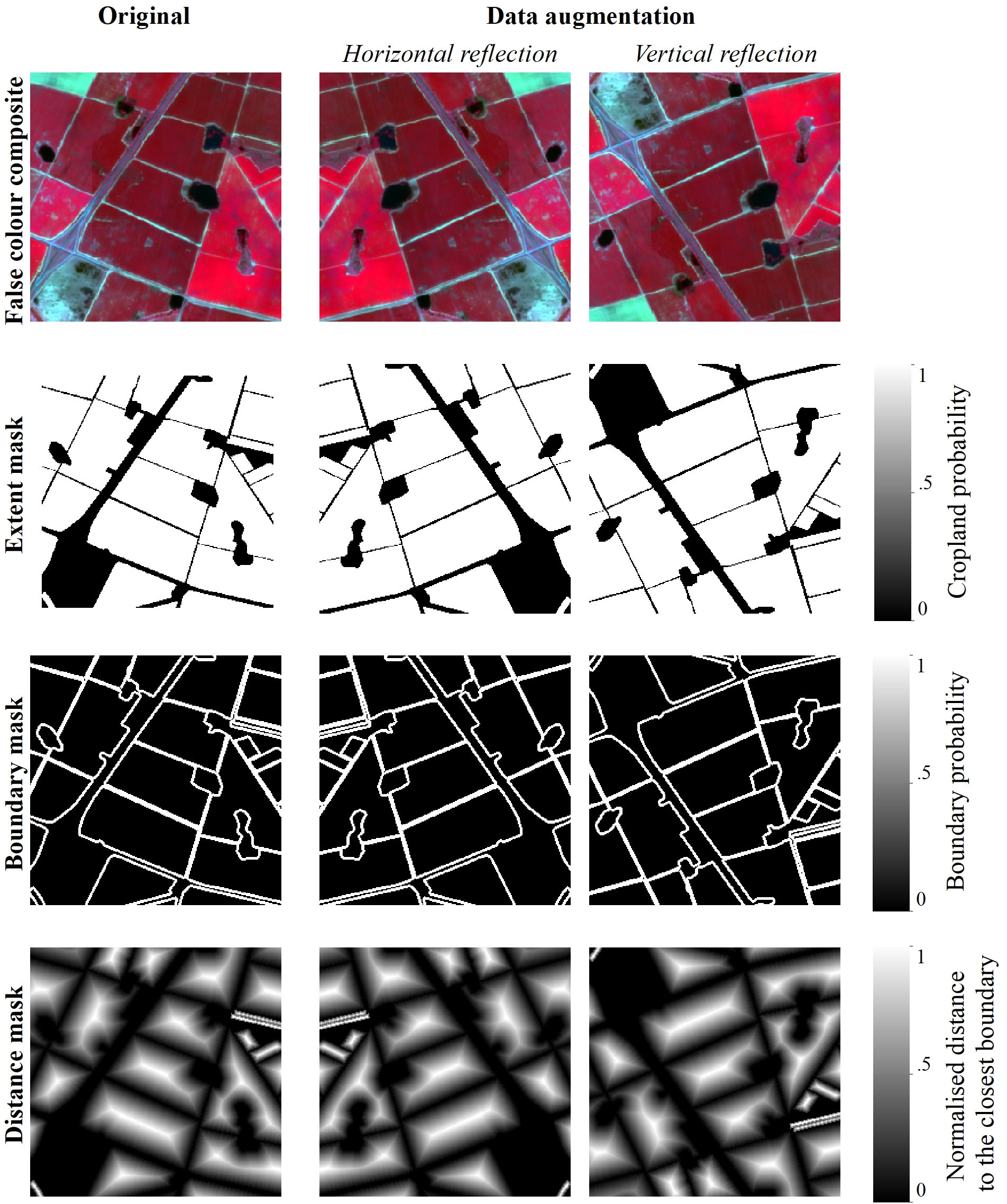}
\caption{\label{fig:data_augmentation} Example of data augmentation for an input image of South Africa (256$\times$256 pixels; 27\textdegree16'S, 27\textdegree28'E). The four rows represent the four inputs to the deep neural network: the satellite image, the extent mask, the boundary mask, and the distance mask. Feeding the network with random transformations of the original data is expected to improve its ability to generalise. }
\end{figure}

 While convolutional neural networks and specifically UNets can integrate spatial information, they are not equivariant to transformations such as scale and rotation~\citep{goodfellow2016deep}. 
Data augmentation  augments the  variance  of  training data, which confers invariance on the network for certain transformations and boosts its ability to generalise. To this end, we flipped  the original images (horizontal and vertical reflections; Fig.~\ref{fig:data_augmentation}) and randomly modified their brightness.  As the size of the training data set was sufficiently large to cover significant variance,  we avoided random rotations and zoom in/out augmentations with reflect-padding because these may break field symmetry, \eg for fields under pivot irrigation. 

\subsubsection{Training}

Most  traditional deep learning methods commonly employ cross-entropy as the loss function for segmentation~\citep{ronneberger2015u}.  However, empirical evidence showed that the Dice loss can outperform cross-entropy for semantic segmentation problems \citep{DBLP:journals/corr/MilletariNA16, DBLP:journals/corr/NovikovMLHWB17}. Here, we relied on the Tanimoto distance with  complements (a variant of the Dice loss), which achieves faster training convergence than other loss functions irrespective of the weights initialisation and keeps a good balance across multiple tasks ~\citep{diakogiannis2019resunet}. Our loss function was defined as the average of the Tanimoto distance  $T(\mathbf{p}, \mathbf{l})$ and its complement $T(\mathbf{1}-\mathbf{p}, \mathbf{1}-\mathbf{l})$: 
\begin{equation}
\label{Tanimoto_wth_dual}
\tilde{T}(\mathbf{1}-\mathbf{p}, \mathbf{1}-\mathbf{l}) = \frac{T(\mathbf{p},\mathbf{l}) + T(\mathbf{1}-\mathbf{p}, \mathbf{1}-\mathbf{l})}{2}. 
\end{equation}
\noindent with
\begin{equation}
T(\mathbf{p}, \mathbf{l}) = \frac{\sum_i p_i l_i}{\sum_i (p_i^2 +  l_i^2) - \sum_i (p_i l_i)} 
\label{eqn:tanimoto}
\end{equation}
\noindent where $\mathbf{p} \equiv p_i \in [0,1]$, represents the vector of probabilities for the $i$th pixel, and $l_i$ are the corresponding ground truth labels. For multiple segmentation tasks, the complete loss function was defined as the average of the loss of all tasks:
\begin{equation}
\begin{aligned}
\tilde{T}_\text{MTL}(\mathbf{p}, \mathbf{l}) = & 0.25 \bigl[\tilde{T}_\text{extent}(\mathbf{p}, \mathbf{l})  + \tilde{T}_\text{boundary}(\mathbf{p}, \mathbf{l}) + \\
 & \tilde{T}_\text{distance}(\mathbf{p}, \mathbf{l}) + \tilde{T}_\text{reconstruction}(\mathbf{p}, \mathbf{l})\bigr]
\end{aligned}
\label{eqn:tanimoto_all}
\end{equation}

\subsubsection{Inference}
Once trained, the model can be applied to any a 256$\times$256 image by a simple forward pass through the network. As convolutional neural networks use contextual information for prediction, it is likely that their classification accuracy varies depending on the location of objects in the input image.  We expect that objects at the centre of input images are well classified because their context is evenly described in every direction. Objects near the edge of input images, however, are missing parts of their context and are thus likely to be misclassified. To mitigate this effect, we created 16 sets of input images by using a moving window of size 256 with stride 64 in every direction, effectively varying the position of image objects in the input images. Inference was performed on all sets of input images which  were averaged using the arithmetic mean.

\subsection{Extraction of individual fields}
\label{sec:postprocessing}

While \resuneta can learn to identify strictly field boundaries, it does not necessarily generate closed boundaries, which is required to extract individual fields. We propose two post-processing methods, one based on thresholding (hereafter the \textit{cutoff} method) and the other based on watershed segmentation (hereafter the \textit{watershed} method), to retrieve closed boundaries and thus,  to achieve instance segmentation and retrieve individual fields. Both leverage multiple semantic segmentation outputs and are data-driven so that they can be automatically optimised using a small sample of reference data. 

\subsubsection{Post-processing methods for instance segmentation}

\begin{figure}[h!!!]
  \centering
  \includegraphics[width=0.99\linewidth]{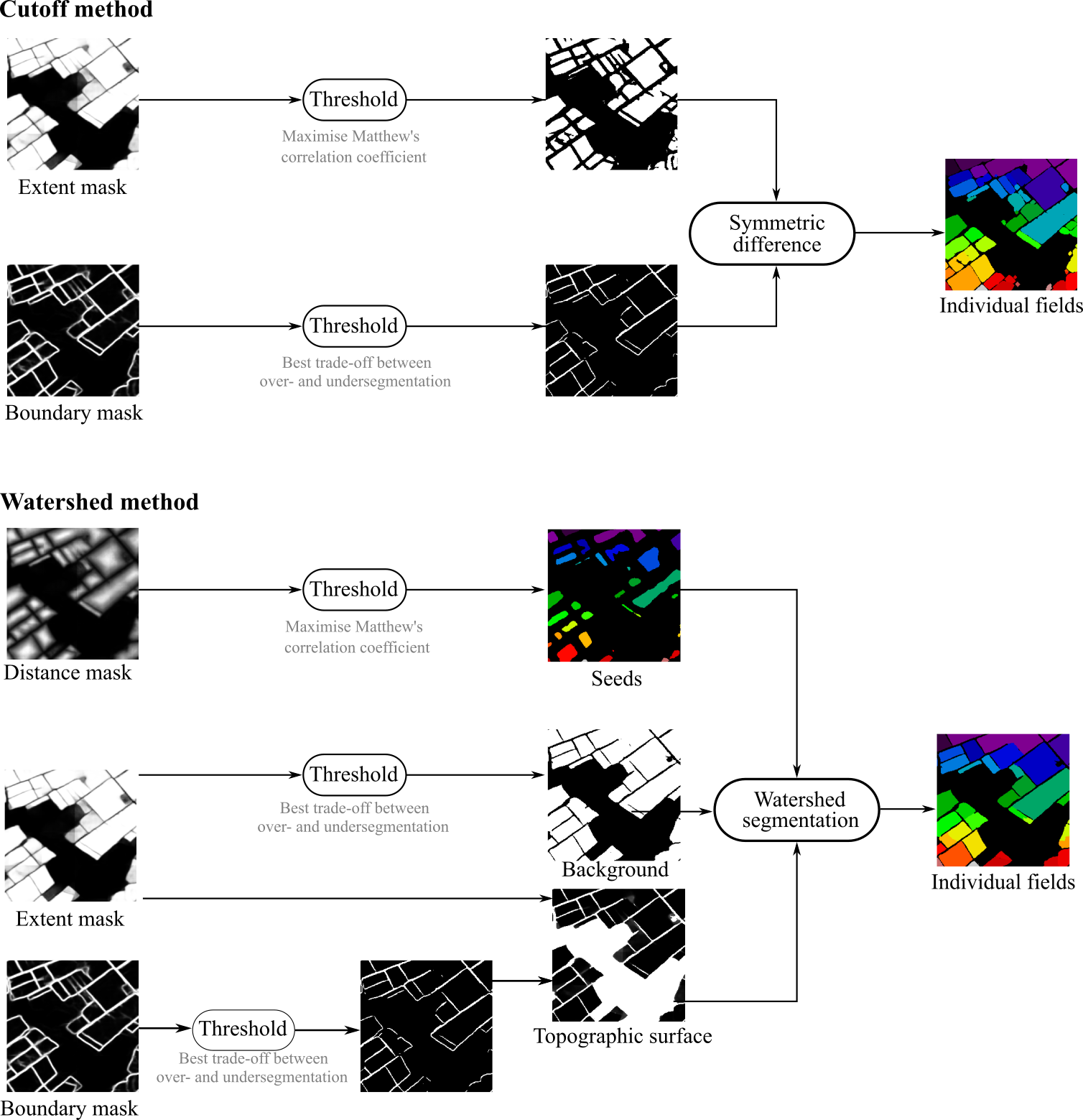}
  \caption{\label{fig:postprocessing} Post-processing methods to generate closed boundaries and extract individual fields. The \textit{cutoff} method generates individual fields  by thresholding and combining the boundary mask and the extent mask. The \textit{watershed} method extracts fields by all segmentation masks in a watershed  algorithm. Seeds for the watershed algorithm are extracted from the distance mask.}
\end{figure}

The \textit{cutoff} method delineates individual fields by thresholding the extent mask and the boundary mask (Fig.~\ref{fig:postprocessing}). A threshold on the extent map defines the extent of fields and, by extension, boundaries between ``field'' and ``non-field'' objects. A threshold on the boundary mask helps further define boundaries between adjacent fields. The symmetric difference of these two binary layers (the set of pixels which are in either of the layers but not in their intersection) produces individual fields with closed boundaries. \\

The \textit{watershed} method  delineates individual fields by applying a  seeded watershed segmentation algorithm on the three output masks. Seeded watershed segmentation considers an input image as a topographic surface (where high pixel values mean high elevation) and simulates its flooding from specific seed points~\citep{meyer1990morphological}, thus generating different growing catchment basins. Catchment basins stop growing when they hit other catchment basins; the lines that separate adjacent catchment basins are their boundaries. By analogy, we can identify the centres of each field (seeds) and grow them until they hit other field boundaries or reach pixels that do not belong to  fields (background). Seeds are defined by thresholding the distance mask and the topographic surface is defined by thresholding the extent mask (Fig.~\ref{fig:postprocessing}). The background, \ie part of the image that does not belong to any field, is defined by thresholding the boundary mask. Unlike the cutoff method that requires two thresholds, three thresholds must be set for the watershed method, one per segmentation mask.

%The third post-processing method, hereafter referred to as the \textit{double watershed} method, seeks to first close all boundaries using a watershed segmentation and second, to define individual fields with a second watershed segmentation (Fig.~\ref{subfig:double_watershed}). First, three regions were defined in the boundary mask based on two thresholds: a foreground region where boundaries are identified with confidence, a background region where absence of boundaries is highly likely, and a region of uncertainty in between. Seeds were defined by the foreground region and were grown based on the distance masks. No fields were identified in the background region. Seeds for the second segmentation algorithm were defined as the inverse of the first segmentation output, \ie seeds are not boundary pixels. Seeds with pixels values smaller than the third thresholds were masks, \ie seeds cannot be outside the cropland extent.  Finally, the second watershed algorithm was instantiated with the extent mask and the newly defined seeds. Again, threshold for the extent mask was defined by maximising the accuracy of the cropland map while the other two were provided the best trade-off between over and undersegmentation.\\

\subsubsection{Threshold optimisation}

The thresholds of the post-processing methods are instantiated using a two-step optimising process: the first step optimises the extent of fields that are extracted (threshold on the extent mask); the second step optimises the fields' shape and size (thresholds on the boundary and distance masks). This procedure thus requires that reference data are available to identify to optimal threshold values (see Section~\ref{sec:data_ref}).\\

First, the threshold for the extent mask is defined by maximising the  Matthew's correlation coefficient~\citep[MCC;][]{matthews1975comparison} between the thresholded and a reference map of the fields' extent (cropland map). The MCC is computed as follows:

\begin{equation}
  \text{MCC} = \frac{\text{TP} ~ \text{TN} -  \text{FP}~ \text{FN}}{ \sqrt{(\text{TP}+ \text{FN})(\text{TP} +\text{FP})(\text{TN}+\text{FP})(\text{TN}+\text{FN})}}   
\end{equation}

\noindent where TP, TN, FP and FN are the true positive rate, the true negative rate, the false positive rate and the false negative rate. The MCC varies between -1 (perfect disagreement) and +1 (perfect agreement) while 0 indicates an accuracy no different from chance. As MCC uses all four cells of a binary confusion matrix, it  is a robust indicator when data are unbalanced~\citep{boughorbel2017optimal}, which is typically the case for boundary detection. \\

Second, the thresholds of the boundary and distance masks are jointly optimised to minimise incorrect subdivision of larger objects into smaller ones (oversegmentation) and an incorrect consolidation of small adjacent objects into larger ones (undersegmentation). The oversegmentation rate ($S_\text{over}$) and undersegmentation rate ($S_\text{under}$) were computed from the reference fields ($T$) and the extracted fields ($E$) data using the following formulae \citep{persello2010novel}:

\begin{equation}
S_{\text{over}} =  1 - \frac{\left|T_i \cap E_j\right|}{\left|T_i\right|} 
\end{equation}

\begin{equation}
 S_{\text{under}} =   1 - \frac{\left|T_i \cap E_j\right|}{\left|E_j\right|}               
\end{equation}

\noindent where $\left| \cdot \right|$ is an operator that calculates the area of a field from $E$ or $T$ and $\cap$ is the intersection operator. These metrics provide rate values ranging from 0 to 1: the closer to 1, the less prone to over- and undersegmentation. If an extracted field $E_j$ intersects with more than one field in $T$,  the location error, and the over- and undersegmentation rates are weighted by  the corresponding intersection areas.\\

\begin{figure}[h!!!]
\centering
\includegraphics[width=0.31\linewidth]{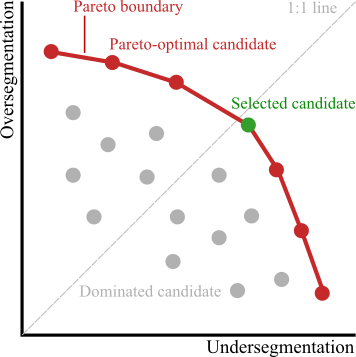}
\caption{\label{fig:pareto_optimal}Identification of the optimal threshold among the Pareto-optimal candidates for instance segmentation. The optimal threshold provides the best balance between undersegmentation and oversegmentation.}
\end{figure}

The average over- and undersegmentation rates are tuned using a multi-objective optimisation procedure that first identified all  Pareto-optimal~\citep{coello2000updated} candidates among those generated. The Pareto-optimal candidates are those candidates for which it is impossible to improve oversegmentation without deteriorating undersegmentation, and \textit{vice versa}.  Finally, the optimal thresholds are given by the Pareto-optimal candidate providing the best trade-off between over and under-segmentation, \ie the closest to the 1:1 line (Fig.~\ref{fig:pareto_optimal}). \\

\section{Data and study sites}
\label{sec:data}

\subsection{Study sites}

\begin{figure*}[h!!!]
    \centering
    \includegraphics[width=\linewidth]{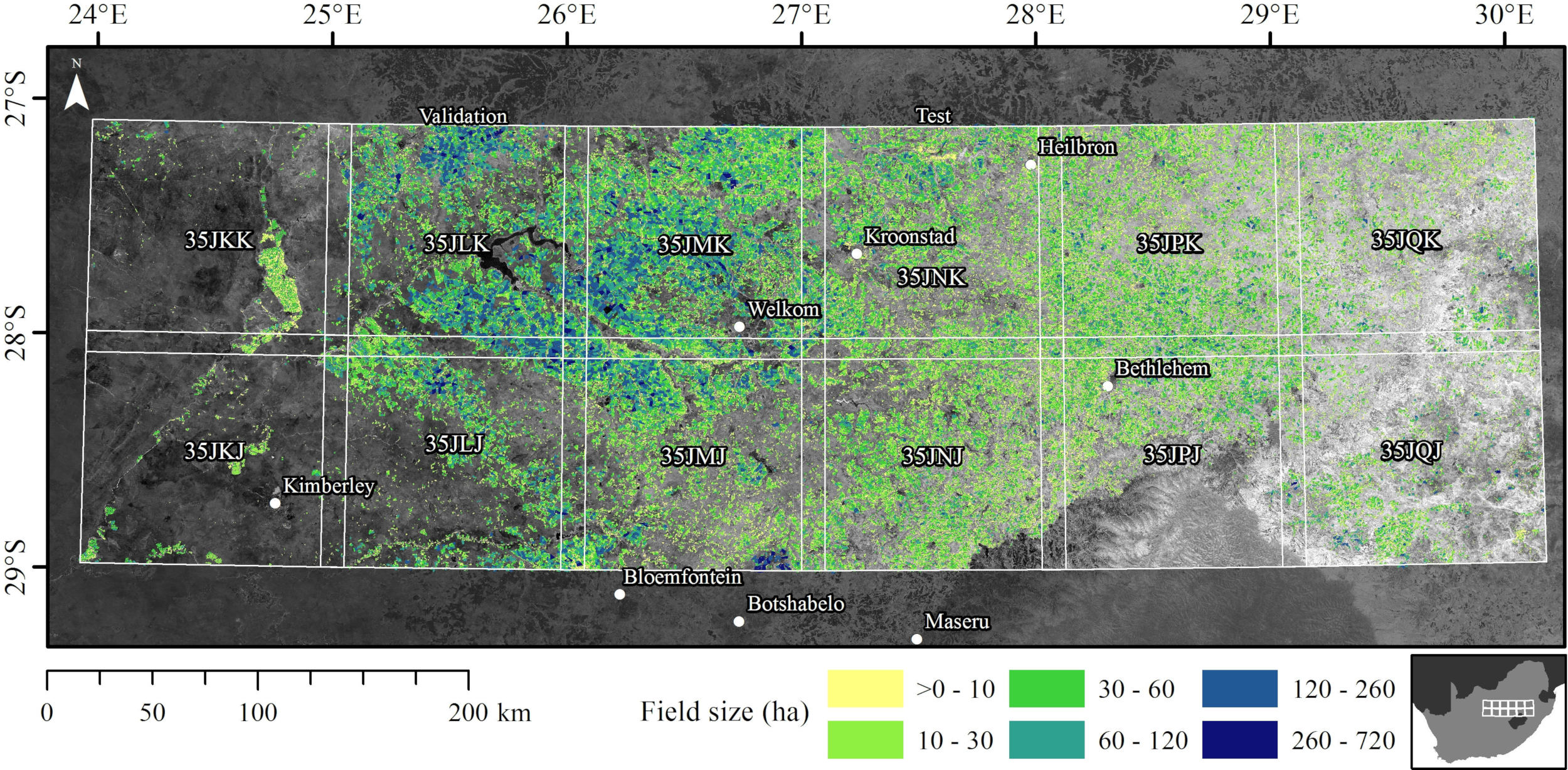}
    \caption{Location of the 208,667 fields available for training and validation in South Africa, the main study site. The field colour indicates the size. Fields in 35JLK were used for validation, and 35JNK for testing. Fields from other tiles were used for training. }
    \label{fig:aoi}
\end{figure*}

Our experiments covered one main study and five secondary sites, all part of the Joint Experiment for Crop Assessment and Monitoring (JECAM\footnote{\url{www.jecam.org}}) network. The purpose of JECAM is to compare data and methods for crop monitoring, with the aim of establishing best practices for different agricultural systems. By selecting JECAM sites, we sought to facilitate future method benchmarking exercises. The main study area encompasses 120,000 km$^2$ of South Africa's ``Maize quadrangle'', which spans across the major maize production area. The field size averages 17 ha and ranges from 1 ha to 830 ha (Fig.~\ref{fig:aoi}). The five secondary study sites of 10,000 km$^2$ each are  in Argentina, Australia, Canada, Russia, and Ukraine (Fig. S1). They were selected to sample  broad-acre cropping systems covering a variety of climate regions, field size, crop types and crop calendar (Table~\ref{tab:site}).  \\

\begin{table}[h!!!]
\centering
\footnotesize
\caption{\label{tab:site}Description of the main and secondary sites.}
\begin{tabular}{L{4.5cm}L{4.cm}}
\hline
\textbf{Main characteristics}                    & \textbf{Main crops and crop calendar}                                \\ \hline

\multicolumn{2}{l}{\textit{South Africa (main site; 120,000 km$^2$) -- 28\textdegree S, 27\textdegree E}}                                                                            \\\vspace{-0.7cm}
\begin{itemize}[leftmargin=*]
  \item Sub-humid to semi-arid climate
  \item Average field size of 17 ha
  \item Flat undulating plains to mountainous
  \end{itemize}                                           
&   
\begin{itemize}[leftmargin=*]
  \item Maize, wheat, sunflower and soybean with some irrigation (centre pivot) 
  \item From December to June; from April to November
  \end{itemize}                                                      
\\

\textit{Argentina (10,000 km$^2$) -- 34\textdegree32'S, 59\textdegree 06'W}                               &                                                                      \\\vspace{-0.4cm}
\begin{itemize}[leftmargin=*]
  \item Temperate humid climate
  \item Average field size of 20 ha (with high variability)
  \item Gentle slopes  
  \end{itemize}                                           
&   
\begin{itemize}[leftmargin=*]
  \item Soybean, maize and wheat (mostly rainfed)    
  \item From June to December and from October to March/April
  \end{itemize}                                                      
\\

\textit{Australia (10,000 km$^2$) -- 36\textdegree 12'S, 143\textdegree 33'E}                               &                                                                      \\\vspace{-0.3cm}
\begin{itemize}[leftmargin=*]
  \item Temperate climate with winter-dominant rainfalls
  \item Fields as large as 250 ha
  \item Gently undulating landscape  
  \end{itemize}                                           
&   
\begin{itemize}[leftmargin=*]
  \item Wheat, barley, canola, and legume crops
  \item From April to December   
  \end{itemize}                                                      
\\

\textit{Canada (10,000 km$^2$) -- 50\textdegree 54'N, 97\textdegree 45'W}                                  &                                                                      \\ \vspace{-0.7cm}
\begin{itemize}[leftmargin=*]
  \item Humid continental climate  
  \item Average field size of 64 ha  
  \item Flat 
  \end{itemize}                                           
&   
\begin{itemize}[leftmargin=*]
  \item Canola, wheat, soybean, maize  
  \item From April to August and May to September/October 
  \end{itemize}                                                                   \\

\textit{Russia (10,000 km$^2$) -- 45\textdegree 98'N, 42\textdegree 99'E}                                  &                                                                      \\\vspace{-0.7cm}
\begin{itemize}[leftmargin=*]
  \item Arid  to humid continental climate  
  \item Fields ranging from 30 to 130 ha
  \item Mostly flat 
  \end{itemize}                                           
&   
\begin{itemize}[leftmargin=*]
  \item Wheat, barley, peas, soybean, sunflower and rapeseed  
  \item From April to October; From September to July
  \end{itemize}                                                      
\\
\textit{Ukraine (10,000 km$^2$) -- 50\textdegree 51'N, 29\textdegree 96'E}                                 &                                                                      \\\vspace{-0.7cm}
\begin{itemize}[leftmargin=*]
  \item Humid continental climate
  \item Fields ranging from 30 to 250 ha
  \item Mostly flat with some hills  
  \end{itemize}                                           
&   
\begin{itemize}[leftmargin=*]
  \item Wheat, sunflower, maize, barley, potatoes, and soybean 
  \item From September to July; from April to October 
  \end{itemize}                                                      
\\ 
\hline
\end{tabular}
\end{table}

\subsection{Satellite data and preprocessing}

%The ability to detect a field boundary  depends on the presence of a contrast at each boundary in one or more spectral bands. And contrast is itself a function of the acquisition date or the temporal feature of the input data and its spatial resolution.

Twelve Sentinel-2 tiles cover the main study site and each secondary site was defined by the extent of a single Sentinel-2 tile (Table~\ref{tab:data}). We chose Sentinel-2 imagery for its 5-day global revisit frequency, which increases the likelihood of cloud-free image acquisition and  provides consistent multi-temporal composites, and  for its 10-m resolution, which is sufficient to resolve most fields in South Africa~\citep{waldner2018local}. Sentinel-2 images are made available free of charge, which facilitates large-scale applications. \\

\begin{table*}[h!!!]
\centering
\footnotesize
\caption{\label{tab:data}Summary of the images used in every experiment of this study. The coordinates indicate the centroid of each site.}
\begin{tabular}{llL{8cm}L{6cm}}
\hline
\textbf{Sensor}                            & \textbf{Site}             & \textbf{Tiles} & \textbf{Acquisition dates/window} \\ \hline

\multicolumn{4}{l}{\textit{4.1. and 4.2. Baseline and comparison with edge filters}} \\

Sentinel-2                   & South Africa                                             &       T35JKJ, T35JKK, T35JLJ, T35JLK, T35JMJ, T35JMK, T35JNJ, T35JNK, T35JPJ, T35JPK, T35JQJ, and T35JQK         &  All images in  03/2017            \\
 & & & \\
\multicolumn{4}{l}{\textit{4.3.1. Generalisation to single-date imagery:} Can the model be applied to single-date images?} \\
Sentinel-2                                         &   South Africa & T35JNK              &  13/03/2017           \\
 & & & \\
\multicolumn{4}{l}{\textit{4.3.2. Generalisation across resolutions and sensors:} Can the model be applied to a 30-m Sentinel-2 image and to a Landsat-8 image?} \\

Sentinel-2       &   South Africa & T35JNK              &  13/03/2017           \\

Landsat-8         &    South Africa      &   170-079             &    29/03/2017          \\ 
 & & & \\

\multicolumn{4}{l}{\textit{4.3.3. Space-time generalisation:} Can the model be applied to other locations and acquisition dates?} \\

Sentinel-2    & Argentina   & T21HTB             & 09/01, 03/02, 10/03, 20/03, 14/04, and 23/07/2018             \\

        & Australia           &  T54HXE & 18/07, 02/08, 11/09, 26/09, 01/10, and 11/10/2018            \\

       & Canada           & T14UNA                 & 26/04, 16/05, 10/06, 15/07, 09/08, and 23/10/2018             \\

        & South Africa  & T35JNK & 13/11 and 03/12/2016; 02/01, 02/04, 01/06, and 11/06/2017              \\
      
             & Russia          &  T35UPR                 & 08/04, 02/06, 22/06, 27/07, and 11/08/2018             \\

        & Ukraine          &  T37TGL                 & 21/04, 11/05, 05/07, 24/08, 13/10, and 18/10/2018             \\

 \hline
\end{tabular}
\end{table*}

Sentinel-2 images were obtained from the European Space Agency's Scientific Hub and were converted to surface reflectance using the Multisensor Atmospheric Correction and Cloud Screening ~\citep[MACCS;][]{hagolle2010multi} algorithm available in the  Sen2-Agri toolbox~\citep{defourny2019near}. In the main site, the same toolbox was used to generate a monthly cloud-free composite of March 2017, which corresponds to the middle of the growing season. Across the images available for compositing, 93\% of the pixels were cloud-free. Five or six cloud-free Sentinel-2 images were processed in each secondary site (Table~\ref{tab:data}). We also sourced the closest  cloud-free  Landsat-8 image to the Sentinel-2 compositing window (Table~\ref{tab:data}).\\

There were only two preprocessing steps: subsetting the blue, green, red, and near infrared bands from the satellite images; and  standardising pixel values so that each band had a mean of zero and a standard deviation equal to one. Standardisation is necessary because neural networks are sensitive to the scale of the input variables and because gradient optimisation methods converge more rapidly with when features have zero mean and unit variance. Standardisation values were obtained from the monthly composite image only, not from images from the secondary sites.
% Images in secondary sites were standardised based on the values obtained in South Africa for the monthly composite.

\subsection{Reference and ancillary data}
\label{sec:data_ref}

For the main site, we obtained boundaries for every field in the area of interest from the South African Department of Agriculture, Forestry and Fisheries \citep{SAfieldboundary}. These were created by manually digitising all fields throughout the country based on 2.5 m resolution, pan-merged SPOT imagery acquired between 2015 and 2017. Field polygons were rasterised at 10 m to match Sentinel-2's grid, providing a wall-to-wall validation data set rich of $>$ 2 billion reference pixels. %(Fig.~\ref{fig:aoi}).
We created three reference layers: a binary layer of the extent of fields, a binary layer of the field boundaries (a 10-m buffer was applied), and a continuous layer representing, for every  within-field pixel, the distance to the closest boundary. Distances were normalised per field so that the largest distance was always one. ``Non-field'' pixels were set to zero.\\

We randomly split the twelve Sentinel-2 tiles into a training (10 tiles), a validation (1 tile, 35JLK) and a test (1 tile, 35JNK) set. Because it is not possible to process entire Sentinel-2 images at once due to limited GPU memory, each Sentinel-2 tile was partitioned into a set of smaller images (hereafter referred to as input images) of a size of 256$\times$256 pixels. Input images on the border of tiles, \ie those with missing values, were discarded in further analyses. This yielded an average of 6150 input images per tile. Input images in the training set were used to train the deep neural network, those in the validation set were used for parameter tuning, and those in the test set were  utilised only for final accuracy assessment. Discrepancies between field boundaries and the Sentinel-2 images were  not necessarily concomitantly acquired. While these discrepancies might be handled during the training phase, their impact is far greater for accuracy assessment. Therefore, we followed a random sampling procedure to visually confirm 1000 field boundaries.\\

For the secondary sites, we first created the layer of the fields' extent for each site by intersecting two global 30-m cropland maps: Globeland 30~\citep{chen2015global} and Global Food Security-support Analysis Data (GFSAD) Cropland Extent~\citep{phalke2017NASA, massey2017NASA, zhong2017NASA}. We resampled this intersection map to 10 m and used information about water bodies, roads, and railways from OpenStreetMap to further mask out pixels wrongly labelled as cropland. Two hundred randomly-selected fields were then manually digitised, for each secondary site, based on Sentinel-2 images and Google Earth images. 

%As the resolution was higher, there might be a penalty.

%\subsection{Field definition}
%None of the literature reviewed clearly defined an agricultural field. The definition of a crop field may vary by disciplines and research questions. For some, a field may represent land tenure, which is related to rules of property rights allocation rather than how the plot of land is managed. But tenure is more closely associated to farm size rather than field size.
%"Field boundaries can be defined as three-dimensional man-made physical barriers, usually intended to prevent livestock from moving from one area to another and/or used to mark ownership boundaries. They include a variety of landscape features, e.g. hedgerows, stone walls, grassy strips, lines of trees and shrubs and any combination of these elements."\citep{petit2003field}

\section{Methods}
\label{sec:experiment}

We conducted several experiments to evaluate the ability of our method to extract field boundaries from satellite images. In a first, baseline experiment, we evaluated the accuracy of our approach in the main site using a single monthly composite of Sentinel-2. Our second experiment compared \texttt{ResUNet-a}'s performance to identify field boundaries against that of a conventional edge detection filter. Our last set of experiments was designed to evaluate the generalisation ability of our approach. Specifically, we assessed its ability to generalise to single-date imagery, to coarser resolutions and different sensors, and to other dates and locations. With these experiments, we wish to gather the necessary evidence to propose guidelines to inform large-scale  field boundary extraction with deep learning.

\subsection{Baseline: ResUnet-a D6 vs. ResUnet-a D7}

\subsubsection{Model training}

We trained a series of \texttt{ResUnet-a D6} and   \texttt{D7} models using Adam~\citep{DBLP:journals/corr/KingmaB14} as an optimiser. Adam computes individual adaptive learning rates for different parameters from estimates of first and second moments of the gradients. It was empirically shown that Adam achieves faster convergence than other alternatives~\citep{DBLP:journals/corr/KingmaB14}. We followed the parameter settings  recommended in \citet{DBLP:journals/corr/KingmaB14}.\\

%: alpha=0.001, beta1=0.9, beta2=0.999 and epsilon=10-8 
%Adam has four parameters: the learning rate, which sets the proportion of weights that are updated; the exponential decay rate for the first moment estimates; the exponential decay rate for the second-moment estimates; and epsilon which is a very small number to prevent any division by zero in the implementation.  \hl{what are our defaults?}\\

Two  training approaches  were tested. The weight decay (WD) approach adds a weight decay parameter to the learning rate in order to decrease the neural network weight and bias values by a small amount in each training iteration when they are updated. This technique can speed up convergence and is equivalent to introducing an L2 regularisation term to the loss function. We trained three models for 100 epochs with different weight decay parameters (10$^{-4}$, 10$^{-5}$, 10$^{-6}$). The interactive approach  involves training a neural network for 100 epochs with a given learning rate (10$^{-3}$) and  resuming training  for another 100 epochs with a reduced learning rate at the epoch that reached the highest MCC. This process was repeated twice for  increasingly lower learning rates (10$^{-4}$ and 10$^{-5}$). As a result, eight models were  compared (three weight decay models and an interactive model for both \texttt{ResUnet-a D6} and   \texttt{D7}). Models were trained on a single node on a high-performance computing system, consuming four NVIDIA Tesla P100 GPUs for a maximum seven days, depending on the training approach. \\

Finally, individual fields were extracted using the outputs of the best model. We compared the two post-processing methods and assessed the significance of their differences using paired Wilcoxon tests~\citep{wilcoxon1945individual}, a nonparametric test for paired data that compares the locations of two populations to determine if one is shifted with respect to the other. For the \textit{cutoff} method, the optimal thresholds were identified using grid search (between 0.01 and 0.99 by step of 0.01). Given the large number of possible combinations in the \textit{watershed} method, a random search of 250 threshold combinations was used instead because it scales better~\citep{bergstra2012random}.

%The back-propagation algorithm was employed to minimise the loss function. Back-propagation consists in a forward pass during which an input image is convolved to calculate the loss \hl{rephrase}, and a backward pass during which \hl{xxx}.\\

\subsubsection{Accuracy assessment}

Model accuracy was evaluated with two types of accuracy metrics: pixel-based metrics and object-based metrics. Object-based metrics latter were particularly important to quantify the accuracy of individual fields that were extracted.\\

Pixel-level accuracy was  computed from two global accuracy metrics derived from the error matrix: Matthew's correlation coefficient; and the overall accuracy (OA), which provides the proportion of pixels that were correctly classified:

\begin{equation}
    \text{OA} = \frac{\text{TP}+\text{TN}}{\text{TP}+\text{TN}+\text{FN}+\text{FP}}
\end{equation}

\noindent We also computed the F-score (F) as a class-wise accuracy indicator:

\begin{equation}
\text{F}_k = \frac{2 \times UA_k \times PA_k}{UA_k + PA_k}
\end{equation}

\noindent where $UA_k$ is the user's accuracy of class $k$ ($UA= \frac{TP}{TP + FN}$) and $PA_k$ is the producer's accuracy ($PA= \frac{TP}{TP + FP}$). For experiments other than those involving South African data, the hit rate was computed as the ratio between the number of fields in the validation data set that were successfully detected and the total number of fields in the validation data set. The hit rate was the preferred thematic accuracy metric because the cropland map in these sites, despite our best efforts, was not accurate enough to be considered as validation data. \\

Object-based accuracy was evaluated with four metrics that collectively capture differences in shape, size and shifts in the location of  extracted fields with respect to target or reference fields. The first two (the oversegmentation rate and the undersegmentation rate) were previously introduced and express incorrect subdivision or incorrect consolidation of fields. The third one, the eccentricity factor ($\epsilon$) reflects absolute differences in shape \citep{persello2010novel}: 

\begin{equation}
     \epsilon = \lVert \text{Eccentricity}_{T_i} - \text{Eccentricity}_{E_j} \rVert
 \end{equation}
% A measure of how much an ellipse deviates from a circle, expressed as the ratio of the distance between the center and one focus of an ellipsoid to the length of its semimajor axis.
 \noindent where the eccentricity indicates how much the shape of a field deviates from a circle (Eccentricity = 0). The fourth and last object-based metric, the location shift ($L$),  was computed to indicate the difference between the centroid location of reference and extracted fields~\citep{zhan2005quality}:
 \begin{equation}
     L = \sqrt{(x_{E}-x_{T})^2 + (y_{E} - y_{T})^2}
 \end{equation}
\noindent where $(x_{E}, y_{E})$ and $(x_{T}, y_{T})$ are the centroids of corresponding fields $E$ and $T$. Here, the location shift was expressed in pixels. If a reference field was matched with multiple extracted fields, the location shift was defined as the area-weighted average all location shifts.\\

We evaluated the significance of the difference in accuracy bewteen the two post-processing methods with paired Wilcoxon  signed-rank tests and declared statistical significance for $P$ values $<$ 0.05.

%Ideally, the extracted and digitised field boundaries align perfectly. However, reference fields may have sub-pixel boundary imprecision due to the vector to raster conversion process.  \\

%Compared to reference fields, extracted fields may be imperfect with errors that include the incorrect subdivision of larger objects into smaller ones (termed  over-segmentation) and the consolidation of small adjacent objects into larger ones (termed under-segmentation). Fields can be evaluated based on location (the spatial coherence between a predicted and reference object) and overlap (over- and under-segmentation, shape comparison).  To capture the accuracy of the field extraction at the field level, rather than pixel level, four object-based accuracy measures  were used  \citep{zhan2005quality, persello2010novel, whiteside2014area, yan2014automated}.\\

\subsection{Comparison with conventional edge detection}

We compared the boundary mask of the model against the edges detected with a Scharr filter.  The Scharr filter, which has a better rotation invariance than other oft-used filters, such as the Sobel or the Prewitt operators, identifies edge magnitudes by taking the square root of the sum of the squares of horizontal and vertical edges. Edges were detected in each spectral band  then averaged. We randomly sampled 1,000 boundary and interior pixels from this layer along with the corresponding pixels from the boundary mask. As the edge detection layer indicates a magnitude rather than a probability, we computed pseudoprobabilities by scaling the magnitude values between 0 and 1 based on their 0.05 and 0.95 percentiles. We then compared the pseudoprobabilities of boundary and interior pixels against the boundary probabilities with  paired Wilcoxon signed-rank tests.

\subsection{Generalisation experiments}

The previous experiment set the baseline for field boundary extraction from a Sentinel-2 image composite. Next, we designed a series of experiments to evaluate the model ability to generalise to a cloud-free single-date image close to the compositing window (section~\ref{sec:singledate}), to the composite image resampled to 30 m  and to a Landsat-8 image close to the compositing window (section~\ref{sec:resampled}), to Sentinel-2 images acquired across the season in the main and in the secondary sites (section~\ref{sec:spacetime}). %The significance of the differences in accuracy was assessed with Wilcoxon tests, and statistical significance was declared for $P$ values $<$ 0.05. %With these experiments, we wish to gather empirical evidence to support recommendations for field boundary extraction at scale with deep learning.

\subsubsection{Generalisation to a single-date image}
\label{sec:singledate}

Monthly composites are attractive for training because, by removing pixels contaminated by clouds and shadows, they maximise the use of training data which are costly to develop. For inference, however, they are less practical than single-date images because of their additional preprocessing needs. Therefore, we tested the assumption that, for inference, composites could be replaced by single-date images without significantly reducing  accuracy. We applied the best \texttt{ResUNet-a} to a cloud-free Sentinel-2 image covering the test area in South Africa and evaluated the accuracy obtained at the pixel and field levels.

%\subsubsection{TOA vs. BOA}
%Humans do no need atmospherically corrected images to draw field boundaries. In the same line, an ideal CNN should be able to extract field  boundaries on uncorrected images, \ie digital numbers. Nonetheless, composites are useful for training because they provide a seamless input image. Therefore, here we test the hypothesis that CNN models trained on composite of top-of-canopy data can be transferred to mono-temporal digital number data. The direct implication is that one can fetch any Sentinel-2 image and readily apply the model to it, removing all need of preprocessing.

\subsubsection{Generalisation across resolutions and sensors}
\label{sec:resampled}
We  evaluated the model's ability to generalise to another sensor coarser by applying  it to a single-date Landsat-8 image covering the South African test site. For comparison, we resampled the monthly Sentinel-2 composite  to 30-m to match the grid of the Landsat-8 image using a nearest neighbour algorithm. This algorithm is naive because it neglects the sensor spatial response~\citep{waldner2018local}, so the fields extracted using the resampled image are expected to be more accurate. 

%did not resampled the validation data so pessimistic bias.

\subsubsection{Generalisation across space and time}
\label{sec:spacetime}
Finally, we assessed the model's ability to generalise in space and time by extracting field boundaries in the main study site and in the secondary sites using cloud-free single-date images across the growing season.  Thematic and geometric changes in accuracy were measured over time for each single-date image. \\

We hypothesised that, given the dynamic nature of cropping systems, some dates would be more appropriate than others. Therefore, we also evaluated the benefit of building consensus from multiple dates as a mechanism to prevent loss of accuracy. Consensus is built by averaging the segmentation masks of multiple observation dates. To identify the  number of observations that delivers consistent improvement, consensus was gradually built along the season. For example, consensus for the third acquisition date was obtained by time-averaging the segmentation mask of the first, second and third acquisition dates in the time series. Individual fields were then extracted from the time-averaged  masks.

\section{Results}
\label{sec:results}

\subsection{Model selection}

\begin{figure}[h!!!]
  \centering
  \subfloat[]{\label{subfig:loss}\includegraphics[trim=0 0 0 1.2cm,clip, width=0.7\linewidth]{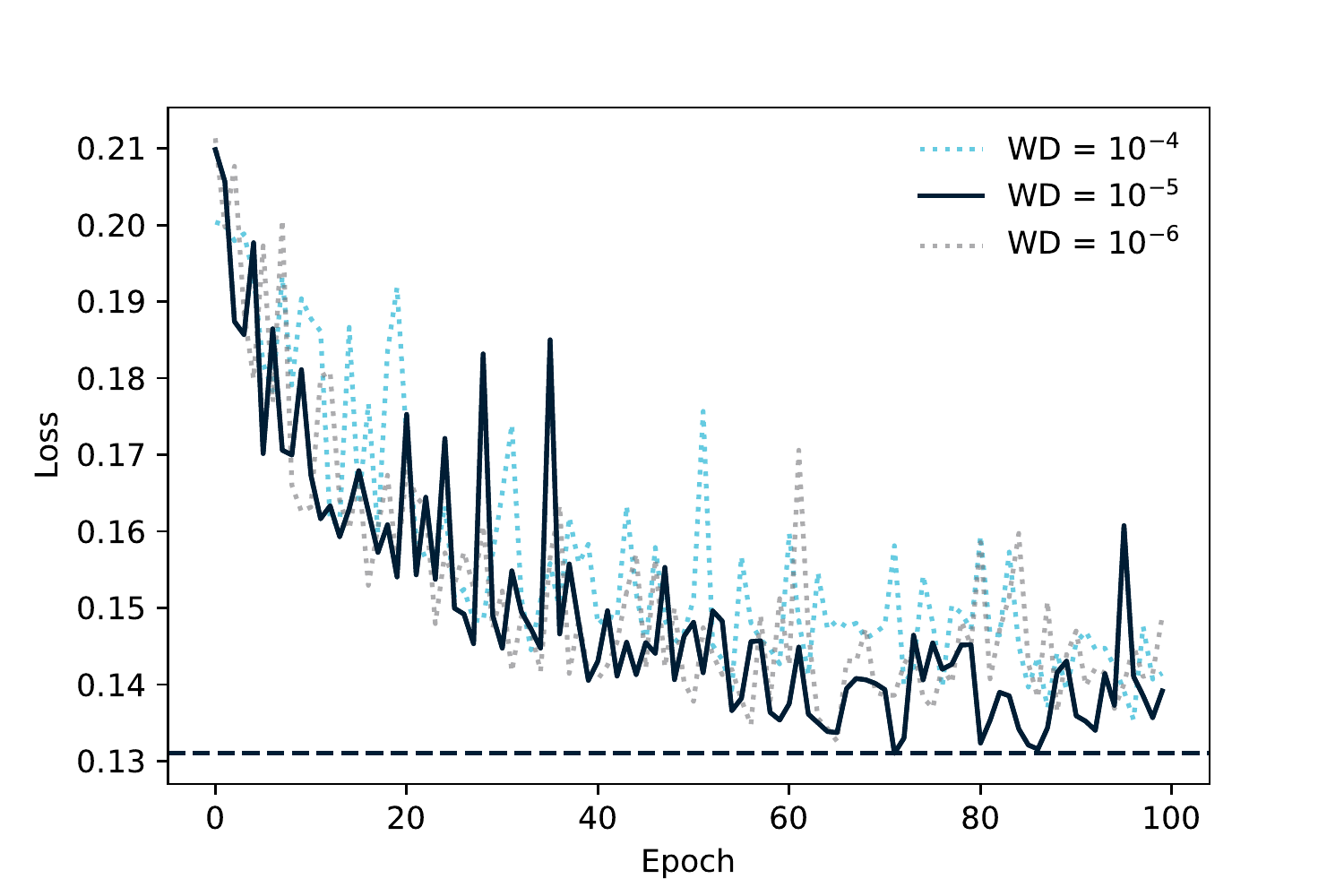}}\\
  %width=0.35\linewidth]{figures/loss_resunet-d6.pdf}}\\
  \subfloat[]{\label{subfig:mcc}\includegraphics[trim=0 0 0 1.2cm,clip, width=0.7\linewidth]{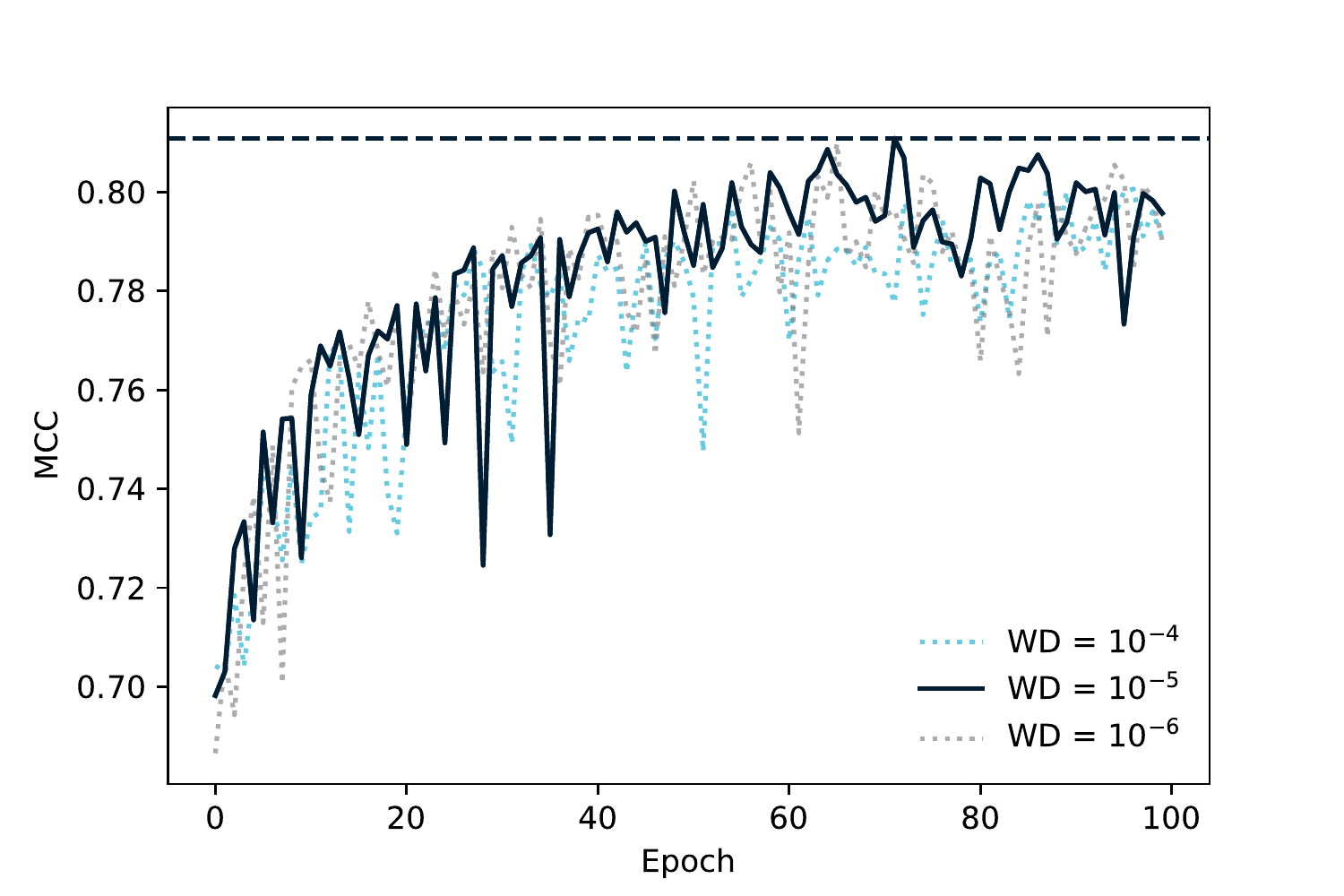}} 
  %width=0.35\linewidth]{figures/mcc_resunet-d6.pdf}} 
  \caption{\label{fig:training} Evolution of the (a) loss function and (b) Matthew's Correlation Coefficient during model training for the procedures involving weight decay (WD). Optimal training (MCC=0.81) was achieved before 80 epochs, then earlier signs of overfitting can be seen.}
\end{figure}

We trained four versions of the \dsix and \texttt{D7} models; the first three versions were parameterised with different weight decay values and, the last version was parameterised interactively. After each epoch, the loss function and the MCC were computed for the test set (see Fig.~\ref{fig:training} for the \texttt{D7} model and the three training modes involving weight decay).\\

Interactive training produced the lowest training loss for both model architectures. However, when validated against the test data, the best \texttt{D6} model was the one trained with highest weight decay performed best (MCC=0.81). We assumed that training the models for more than 100 epochs was unlikely to improve accuracy because the training curves started to show signs of overfitting after 80 epochs. Overall,  \texttt{ResUNet-a D7}, the deeper model, yielded the highest accuracy (MCC=0.82), so it was selected for all subsequent analyses.

\begin{table}[h!!!]
\footnotesize
\centering
\caption{\label{tab:pix_validation} Accuracy of the two model architectures for different training modes. }
\begin{tabular}{llll}
\hline
 \textbf{Training mode} & \textbf{Training loss} &  \textbf{Validation loss} & \textbf{Validation MCC} \\ \hline
 \multicolumn{4}{l}{\textit{ResUNet-a D6}} \\
 WD = 10$^-4$               &  0.1212                      &                        0.1310                      &  0.8109                       \\
 WD = 10$^-5$               &  0.1118                      &                                   0.1326             &    	0.8099                     \\
 WD = 10$^-6$               &   0.1095                     &                                   0.1310              &  	0.8109                      \\\vspace{0.1cm}
 Interactive           &   0.0947                                          &  0.1344  &                           0.8091                    \\ 
\multicolumn{4}{l}{\textit{ResUNet-a D7}} \\
WD 10$^-4$               &  0.1119                      &                         0.1369                     &    	0.8012                      \\
 WD 10$^-5$               &  0.1017                                            &     0.1339                    & 	0.8086                        \\
 WD 10$^-6$               &  0.0979                      &                                0.1326                &    	0.8099                     \\
 Interactive           & \textbf{0.0801}*                       &                             \textbf{0.1285}*                 &  	\textbf{0.8158}*                       \\ \hline 
\end{tabular}
\end{table}

\subsection{Assessment of the baseline model}

We assessed the accuracy of the \dseven model using pixel-based accuracy metrics  (Table~\ref{tab:accuracy_pixels}). The overall accuracy was 92\% and the MCC reached 82\%. The cropland class had a slightly lower F-score (89\%) than the non-cropland class (93\%). Tuning threshold of the extent map to maximise MCC  led to only marginal differences ($<$1\%) compared to using a default threshold of 50\%. Nonetheless, as we sought to achieve the classification with the most balanced accuracy for cropland and non-cropland, we retained the optimised threshold. \\

\begin{table}[h!!!]
\footnotesize
\centering
\caption{\label{tab:accuracy_pixels} Pixel-based  assessment of the extent map of South Africa for the baseline experiment, the generalisation to single-date imagery, and to a resampled 30-m Sentinel-2 image. }
\begin{tabular}{lcccc}
\hline
\textbf{Threshold} & \textbf{OA} & \textbf{MCC} & $\mathbf{F_{C}}$ & $\mathbf{F_{NC}}$ \\ \hline
\multicolumn{5}{l}{\textit{Baseline}}    \\
Default threshold (50\%)     & 91.69       & 82.23        & 89.09                    & 93.29                        \\
Optimised threshold (38\%)   & 91.66       & 82.46        & 89.26                    & 93.18                       \\ 
\multicolumn{5}{l}{\textit{Generalisation to a single-date image}}    \\
Optimised threshold (38\%)   & 0.893      &    0.782     &  0.872     &  0.909                     \\ 
\multicolumn{5}{l}{\textit{Generalisation to a resampled 30-m image}}    \\
Optimised threshold (61\%)   & 0.856      &    0.696     &  0.798     &  0.889                     \\ \hline
\multicolumn{5}{l}{\tiny{OA: Overall accuracy; MCC: Matthew's correlation coefficient}}\\
\multicolumn{5}{l}{\tiny{$F_C$: F-score for the cropland class; $F_{NC}$: F-score for the non-cropland class}}
\end{tabular}
\end{table}

\begin{figure}[h!!!]
\centering
\includegraphics[width=0.99\linewidth]{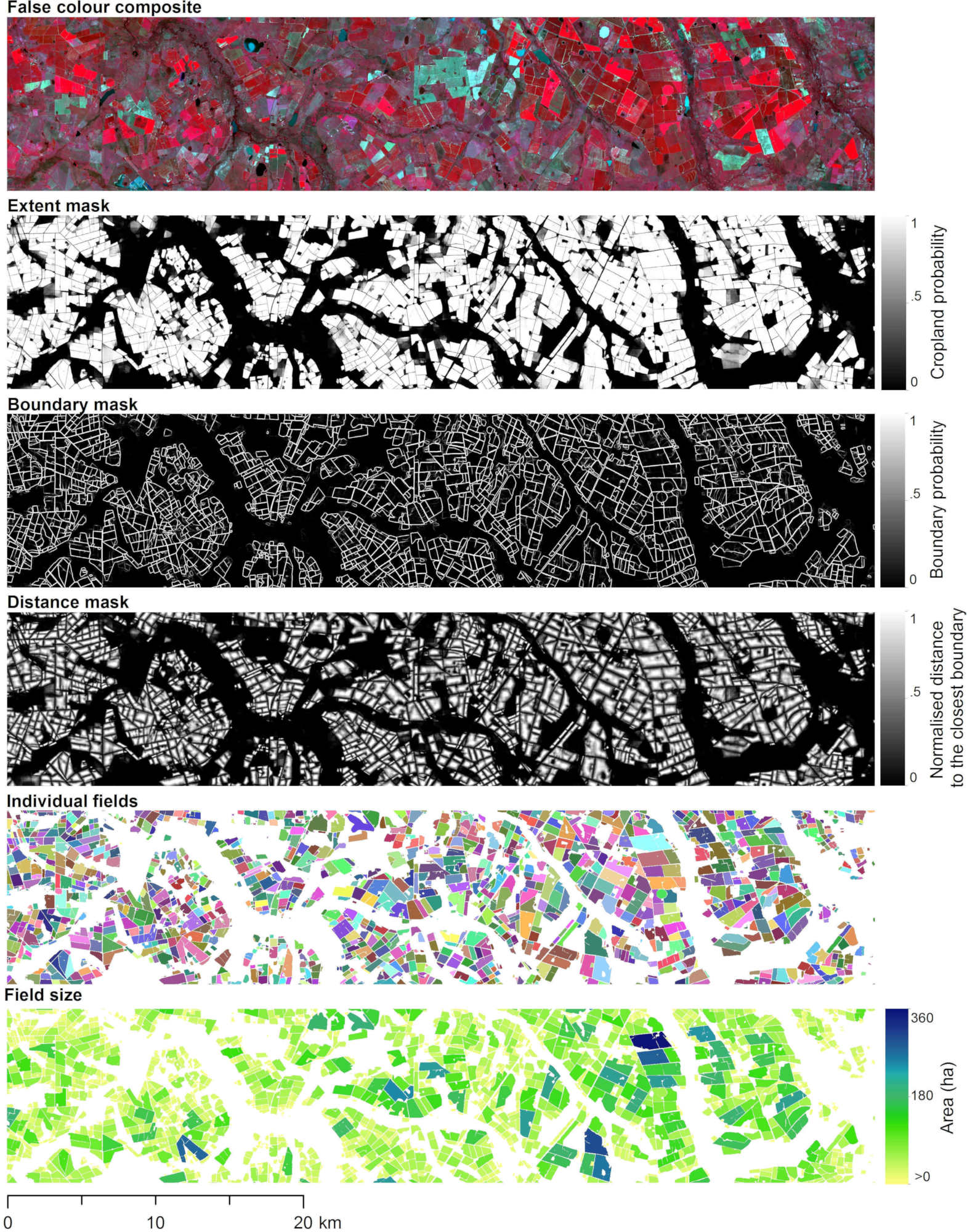}
\caption{\label{fig:resuneta_fields_area} Field extraction examples in  South Africa. Each inset is 77 km $\times$ 40 km centred on 27\textdegree E 45'', 27\textdegree S 33''.}
\end{figure}

We extracted 55,720 fields ranging up to 380 ha (mean = 13 ha) from the Sentinel-2 image of the test region in South Africa (Fig.~\ref{fig:resuneta_fields_area}). Ninety-nine per cent of the reference fields were identified (Table~\ref{tab:all_tab}) with high accuracy for both shape (all metrics $>$ 0.85) and position (location shift of 7 pixels). Despite being simpler, the \textit{cutoff} approach achieved similar results to the \textit{watershed} approach. % If the true difference between two treatment groups is so small that it is clinically irrelevant
This might be explained by the number of possible combinations of parameters to optimise in the \textit{watershed} approach. A more advanced optimisation method, such as the Bayesian optimisation, may be key to find the optimal combination.  We report also no clear benefit in tuning the cutoff values compared to using a default cutoff value of 50\%---the oversegmentation rate was the only metric for which the impact was $>$ 0.05.

% Ldom 
% Sunder 0.814 Sover 0.898 ecc = 0.997 hit rate 0.994

\begin{table*}[h!!!]
\footnotesize
\centering
\caption{\label{tab:all_tab} Object-based assessment for experiment, for the generalisation to single-date imagery, for the generalisation to other resolutions and sensors, and across space and time. The range of over- and undersegmentation, eccentricity, and hit rate is from 0 to 1, with 1. The location shift is given in pixels.}
\begin{tabular}{llllllll}

\hline
\textbf{Image}              & \textbf{Location}             & \textbf{Post-processing} & \textbf{Hit rate} & \textbf{OverSegmentation} & \textbf{Undersegmentation} & \textbf{Eccentricity} & \textbf{Shift} \\ \hline
\multicolumn{8}{l}{\textit{Baseline}}                                                                                                                                                                           \\
Sentinel-2 & South Africa & Naive                    & 0.994              & 0.870                       & 0.869                        & 0.997                  & 7              \\ %70m
                            &                               & Watershed         & 0.995              & 0.870                       & 0.858                        & 0.997                  & 6              \\ %64m
%                            &                               & Cascade Watershed        & 99.5              & 90.0                       & 83.2                        & 99.7                  & 64              \\
\multicolumn{8}{l}{\textit{Generalisation to a single-date image}}                                                                                                                                                          \\
Sentinel-2                  & South Africa                  & Cascade Watershed       & 0.990              & 0.849                       & 0.847                        & 0.998                  & 7              \\
\multicolumn{8}{l}{\textit{Generalisation across resolutions and sensors}}                                                                                                                                          \\
Sentinel-2 30m              & South Africa                  & Naive                    & 0.883              & 0.695                       & 0.696                        & 0.996                  & 5             \\
Landsat-8                   & South Africa                  & Naive                    & 0.791              & 0.665                       & 0.673                        & 0.995                  & 6             \\
\multicolumn{8}{l}{\textit{Generalisation across space and time - Consensus approach}}                                                                                                                                            \\
Sentinel-2                  & Argentina                     & Naive                    & 0.965             & 0.818                      & 0.701                       & 0.910                 & 20              \\
                            & Australia                     & Naive                    & 0.886             & 0.764                      & 0.806                       & 0.788                 & 20              \\
                            & Canada                        & Naive                    & 0.965             & 0.796                      & 0.800                       & 0.819                 & 20              \\
                            & Russia                        & Naive                    & 0.935             & 0.828                      & 0.830                       & 0.955                 & 22              \\
                            & Ukraine                       & Naive                    & 0.980             & 0.864                      & 0.855                       & 0.890                 & 14              \\ \hline
\end{tabular}
\end{table*}

\subsection{Comparison with conventional edge detection}

\begin{figure}[h!!!]
\centering
\includegraphics[width=0.99\linewidth]{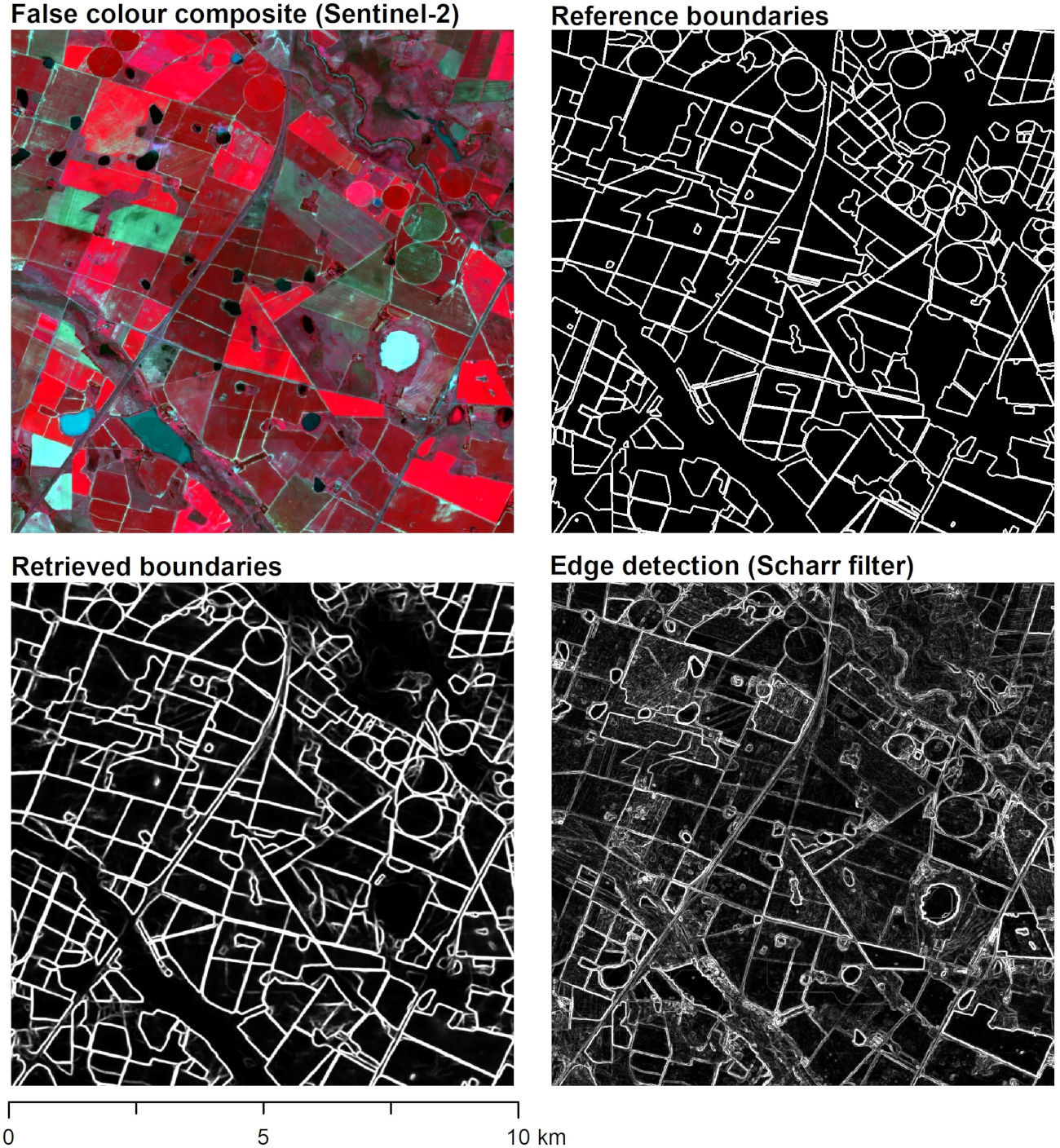}
\caption{\label{fig:resuneta_vs_edge} Comparison  of the boundaries retrieved by our model against references boundaries and edges detected with a Scharr filter for a 10-km $\times$ 10-km region in the test region (27\textdegree E 49'', 27\textdegree S 30'').}
\end{figure}

We compared the boundary mask retrieved by our \resuneta against the edges detected by a Scharr filter (Fig.~\ref{fig:resuneta_vs_edge}). The edge-based method  yielded significantly weaker boundaries ($P <$0.001) and noisier interiors ($P <$0.001). With conventional edge detection, interior pixel values were on average higher and spread across a wider range than those obtained with \texttt{ResUNet-a}. As the neural network learns to be sensitive to certain types of edges, the retrieved edges become clearer.

%Noisy and pixel value are not explicitly tied to  probability of finding a boundary pixel.
%As a result, high pixel values do not systematically denote boundary pixels (shifted). Similarly, interior pixels had higher pixel values and wider spreader

\subsection{Generalisation to single-date imagery}

Feeding the model with single-date data instead of a monthly composite had little effect on its performance (Table~\ref{tab:all_tab}). For instance, the over- and under-segmentation rates dropped by 0.02-0.05 while the offset remained unchanged. These differences were statistically significant only for the undersegmentation rate ($P$ $<$ 0.001), which suggests that extracting field boundaries from single-date images is a viable option.

\subsection{Generalisation across resolutions and sensors}

Resampling to 30 m  reduced the hit rate by only 0.06. This loss of accuracy illustrates the impact of the sensor spatial response (which blurs the image and was neglected during resampling) on the ability to detect cropland~\citep{waldner2018local}. \\% For Landsat-8, we thus estimate that the spatial resolution explained 10\% of the loss of the accuracy, while the sensor spatial response, the difference in spectral response, and the difference in acquisition date collectively explained another 3\%.\\

For individual field extraction,  our results showed that the model was more sensitive to a change of resolution than to a change of sensor (Table~\ref{tab:all_tab}). When extracting field boundaries from 30-m  rather than 10-m Sentinel-2 data, only the hit rate changed significantly (0.88 to 0.79). These changes ought to be related to loss of spatial details and increased difficulty of extracting boundaries, especially in more fragmented landscapes. On average, the  size of the undetected fields (15 ha) was smaller than that of detected fields, which was significant ($P$ $<$ 0.001). As the model achieved similar performances with Landsat-8 data as with 30-m Sentinel data, we concluded that it exhibits good generalisation capability across scales and sensors.

%Talk about differences in spectral bands and date in discussion. \\

%For the 30 m cropland map: optimising threshold improved by less the 2\% the cropland extent.

\subsection{Generalisation across time and space}

\begin{figure*}[h!!!]
\centering
\includegraphics[width=0.89\linewidth]{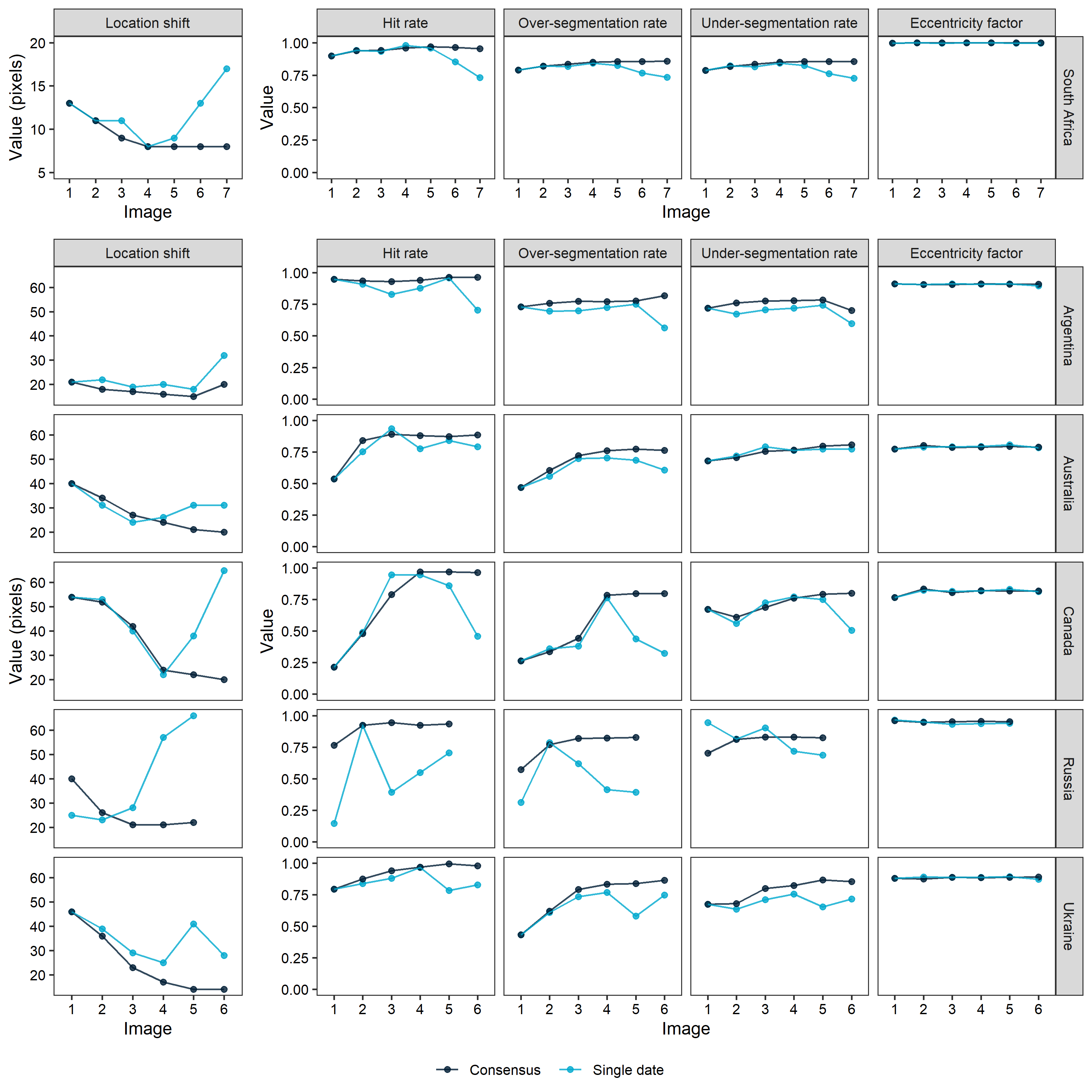}
\caption{\label{fig:singledate_vs_consensus} Model accuracy for space-time generalisation. The $x$-axis represents the image position in the time series (single-date processing) or the number of images that were averaged to build consensus (so that consensus  for image 3 is a time-average of image dates 1,  2 and 3). The model was able to generalise across space and time, but considerable temporal variability was observed. Building consensus is a simple and effective approach to mitigate this variability in accuracy: it was generally at least as good as single-date predictions. Most benefit of the consensus approach was realised with four images. }
\end{figure*}

The acquisition date had a significant impact on the accuracy of the field extraction process in South Africa (Fig~\ref{fig:singledate_vs_consensus}). The hit rate and the location shift were particularly affected, with changes from 0.75 to 0.99 and 7 pixels to 17 pixels, respectively. Building consensus successfully reduced variability: it achieved equal, if not better, performance than single-date cases.  The rate of improvement due to consensus reduced after four images.\\

The impact of changes in acquisition dates was even more marked in secondary sites (Fig.~\ref{fig:singledate_vs_consensus}). In Canada, for instance, the hit rate ranged from 0.21 to 0.97 and the oversegmentation rate from 0.26 to 0.80. Of all the metrics, eccentricity was the least sensitive. Unlike in the main site, it was critical to optimise thresholds in secondary sites, which indicates that local parameter tuning is required for good generalisation. \\

\begin{figure}[h!!!]
\centering
\includegraphics[width=0.95\linewidth]{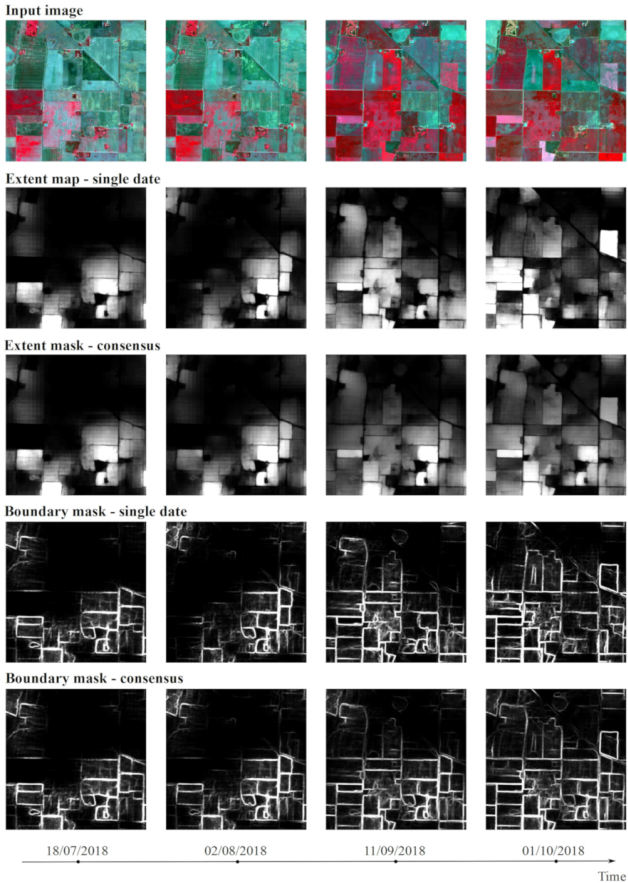}
\caption{\label{fig:consensus} Building consensus for an image subset in Australia. Single-date masks fail to capture all fields and all boundaries but missed patterns can be detected at later in the season. By averaging multiple predictions, the consensus approach is a computationally-cheap option to safeguard against loss of accuracy. }
\end{figure}

\begin{figure*}[h!!!]
\centering
\includegraphics[width=0.88\linewidth]{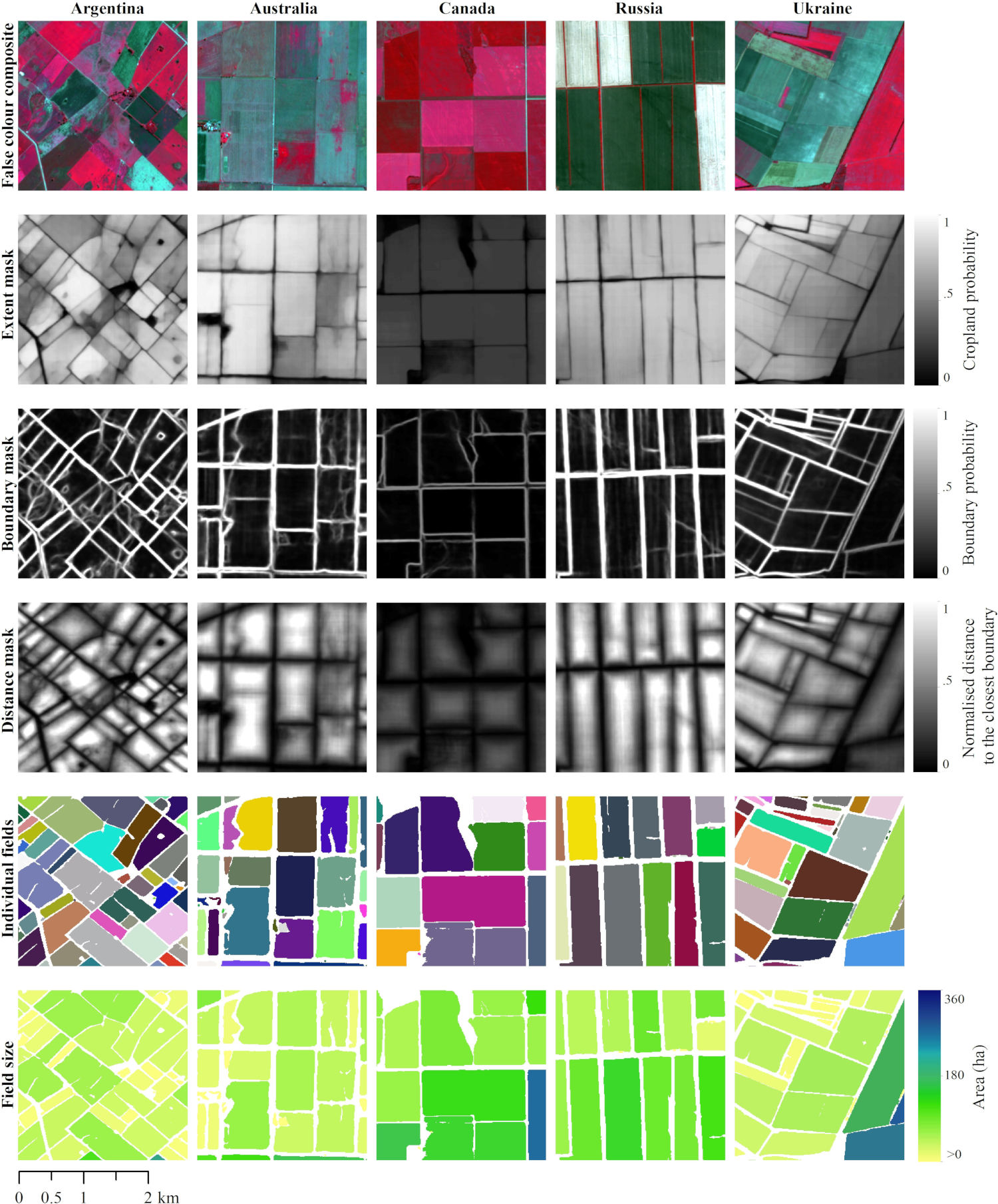}
\caption{\label{fig:allsites}  Field extraction in the secondary sites with the consensus approach: Argentina (34\textdegree32'S, 59\textdegree 06'W), Australia (36\textdegree 12'S, 143\textdegree 33'E), Canada (50\textdegree 54'N, 97\textdegree 45'W), Russia (45\textdegree 98'N, 42\textdegree 99'E), and Ukraine (50\textdegree 51'N, 29\textdegree 96'E).  Each inset represents a  2.5-km $\times$ 2.5-km subset of the full  100-km $\times$ 100-km secondary sites.}
\end{figure*}

While it was possible to yield high accuracy with a single image, the success of the extraction was highly variable, depending on the date of image acquisition. Building consensus was highly effective at reducing the variability in accuracy, which improved the model's ability to generalise across  the board (Figs.~\ref{fig:singledate_vs_consensus} and~\ref{fig:consensus}). For instance, the hit rate after consensus exceeded 0.95 in every site except  Australia (0.86).  The retrieved fields and matched the geometry of reference fields, with under- and oversegmentation rates ranging from 0.7 to 0.86. The benefit of consensus plateaued after combining four images. Fig~\ref{fig:allsites} illustrates the boundary extraction process in all secondary sites.

\section{Discussion} 

\label{sec:discussion}

% 1) On what issue we have to concentrate, discuss or elaborate?
%2) What solutions can be recommended to solve this problem? 
%3) What will be the new, different, and innovative issue? 
%4) How will our study contribute to the solution of this problem

\subsection{Deep learning for field boundary extraction}

We addressed the problem of field boundary extraction from satellite images as multiple, correlated semantic segmentation tasks to be solved by a convolutional neural network called \texttt{ResUNet-a}. The network was trained on a monthly Sentinel-2 composite of South Africa and achieved high accuracy at the pixel and field levels. The same model, without retraining,  performed  accurately when applied to single-date imagery of the same site acquired by Sentinel-2 and Landsat-8.  The ability to extract boundary from a single-date image is cost-effective and a great advantage in areas where cloud cover is persistent such as in the tropics. The model also generalised well in five different cropping systems and for a range of acquisition dates.  Building consensus by time-averaging predictions from multiple single-date images further improved accuracy.  Such generalisation abilities clearly indicate that our approach successfully minimised over-fitting, a prevalent problem in deep learning. It  suggests that convolutional neural networks reduce the importance of spectral and temporal information by heavily exploiting  multilevel contextual information.\\

%on a single composite image: texture is heavily used.  Much better job than conventional edge detection methods. Our approach successfully minimised over-fitting, a prevalent problem in deep learning. As a result, the model generalised well in a range of settings such as... The ability to extract boundary from a single-date images is cost-effective but is a great advantage in areas where cloud cover is persistent such as in the tropics. Yet, there is a gain in accuracy when combining multiple single date predictions.\\

This study shows that addressing the problem of field boundary extraction from satellite images as multiple semantic segmentation tasks with a convolutional neural network achieves excellent performance ($>$85\%) both at the pixel level and the  object level. We believe that this is because our neural network learns to discard edges that are not part of field boundaries and to emphasise those that are using spectral and multilevel contextual features  that are relevant for multiple correlated tasks. As a result, our model identified boundaries with significantly more sharpness and less noise than a benchmark edge-based method. The extent of the cropping area in the main site was mapped with an accuracy of $\sim$90\%,  which is comparable to state-of-the-art methods for cropland classification ~\citep[\eg][]{inglada2017operational,  waldner2017national, oliphant2019mapping}. This is nonetheless remarkable because, while most published methods identify cropland with multitemporal features that capture the dynamics of a pixel across the growing season, our method did so with a single monthly composite. This suggests that convolutional neural networks reduce the importance of temporal information by heavily exploiting hierarchical contextual information. However, reported difficulties in distinguishing certain classes (such as cropland from grassland) from a single image indicate that temporal information still needs to be considered in certain cases. \\   % (\textit{1st move: the performance of our model compared to published results and possible reasons of success})

%In a series of experiments, we evidenced the ability of our model to generalise across space, time, image types, resolutions and sensors. \\

Our convolutional neural network is able to detect fields and their boundaries across different resolutions and different sensors. Convolutional neural networks are designed to be scale-invariant. The features created by the model are object-based: they do not rely on individual pixel values but on all those belong to an object. This mechanism does not depend on a particular resolution. %If a model is trained to recognised fields and field boundaries, it will create feature that emphasize these 
%\Foivos{This is something to be expected, as CNNs are designed to be scale invariant, meaning that the features created to perform a task are object based, and do not rely on individual pixel values. That is, the features created depend on all of the pixels that belong to an object. Putting it differently, a CNN creates features from objects as a whole. This mechanism  does not depend (to some extent) on a particular image resolution: if a CNN is designed to recognize cats, it will create features that emphasize  characteristics of cats (e.g. shape of ears, whiskers, eyes, nose etc). This is irrespective of the input image size. After all, convolution operators can be applied without modification into images of different shapes. This mechanism is not completely fool proof though, changing the resolution (during inference) of an input image can affect performance. However, to the best of our knowledge, there does not exist a theoretical measure for quantifying performance differences. This is something to be deduced experimentally.   
It is also possible that the  ability of \texttt{ResUNet-a}  to derive field boundaries from coarser images can  be related to the  usage of the multiple parallel atrous convolutions that identify features at different scales, but this remains to be verified   experimentally. When we applied 
the model to coarser Sentinel-2 data we observed a reduced  hit rate and a  geometric accuracy reduction by 10-15\%. This was mostly driven by smaller fields not being resolved in the 30-m image. Differences between the resolution of the reference (2.5 m) and extracted (30 m) data might introduce an artificial bias in the accuracy assessment. Applying the model to Landsat-8 data  decreased the geometric accuracy to $\sim$70\% and the hit rate to $\sim$80\%. Differences in the spectral and spatial responses of Sentinel-2 and Landsat-8 partly explain this loss in accuracy. Nonetheless, the average accuracy metrics were relatively similar to those reported for field boundary extraction  using multi-temporal Landsat images across South America~\citep{graesser2017detection}. The  multi-modal capabilities of our approach open up the possibility to operate cross-platform in cloud prone areas and hint at exploiting open image archives imagery, such as the Landsat collection \citep{egorov2019landsat}, to reconstruct temporal patterns of field sizes. Transfer to even coarser ($>$30 m) and higher ($<$10-m) resolutions remains to be empirically evaluated. \\ % (\textit{2nd move: illustrate the possibilities that are enabled by good model performance across resolutions and sensors})

%was relatively fast to train as it converged in less than \hl{xx} epochs.\\

The date of image acquisition matters, especially when extracting fields in previously unseen areas. In all sites, accuracy varied with  acquisition date, with differences as large as 75\%. Part of this sensitivity may be attributed to the training data set that consisted solely of images from a single month. Training the neural network with images spanning the whole season or covering more locations would be likely to reduce this effect. However, high model performance for at least one date per site suggests that changes of local image contrast linked to crop calendars are the main cause of these variations in accuracy. Local knowledge of  crop calendars and cropping practices might dictate which image to select but this is fraught with uncertainty  and depends on data availability.  \\ %(\textit{3rd move: reasons for the temporal variability of accuracy})

Building consensus by averaging predictions over time is a simple, yet effective, solution to consistently improve accuracy. Consensus halved location errors and, in most cases, yielded accuracy at least as large as using single-date images. Although it had been previously suggested that combining multiple image dates improves boundary detection \citep{watkins2019comparison}, ours is the first study to clearly demonstrate the relationship between accuracy and the number of images and to show that predictions from four images well-spread across the growing season are required to achieve most benefits. By covering changes in crop development, the likelihood of capturing an image with sharp contrast increases. Alternatives to harness the temporal dimension include replacing the input image with temporal features~\citep[for instance, see][]{graesser2017detection}  or adding a temporal dimension to the convolution filters. Compared to these alternatives, the consensus approach has lighter processing requirements since most space agencies now directly provide surface reflectance images for download.\\% (\textit{4r move: we demonstrated that building consensus protects again undesirable drops in accuracy})

None of our experiments included transfer learning. Transfer learning adapts a neural network to a new region by freezing the feature extraction layers and fine tuning the last layers based on a  small number of labelled data~\citep[][see also \citealt{7301382}]{Pan:2010:STL:1850483.1850545}. This means that our model might have been exposed to reflectance values that it had never been  previously exposed to. Implementing transfer learning would amount to updating the multitasking head of \texttt{ResUNet-a} and would likely further boost accuracy. The promising generalisation capabilities seem to indicate that requirements in terms of training set size  are relatively low. %(\textit{5th move: Experiment did not include transfer learning, so further accuracy gains are foreseeable})

\subsection{Recommendations for large-scale field boundary extraction}

The overarching objective of this paper was to gather evidence to help lay the blueprints of a data-driven system to derive field boundaries at scale. Based on our results, recommendations can be made as to how to efficiently implement such a system using Sentinel-2 imagery.\\ %(\textit{6th move: our study is so comprehensive that we can formulate guidelines for future work})

\textit{Train on composite images using blocks of continuous reference data.}
Deep learning relies on large amounts of training data, which are usually costly and time-consuming to collect. For field boundary detection, it appeared that the model was able to generalise across a range of locations from a local data set. This illustrates that, if one wishes to collect  new training data sets, sample sizes are reasonable, especially as it can be artificially inflated using data augmentation techniques.  New training data should be collected in blocks  so that the model can learn to identify field boundaries in their context. While one must strive to collect accurate training data, ours were not completely  free of errors but this did not seem to significantly impact the model's performance. Leveraging existing data sets, such as those from the European Land Parcel Identification System, is another option to cut down data collection costs. We also recommend training the model on monthly composites from across the season. Monthly composites are particularly appealing for two reasons: 1) in most cropping systems, they provide consistent, cloud-free observations across large areas, and 2) they minimise the amount of training data wasted because of contamination from clouds and cloud shadows. With Sentinel-2's five-day revisit frequency, several monthly  composites per year should be obtainable in most cropping systems.\\

%Furthermore, to maximise their usage Training data for boundary are most usable when all pixels in an input image are labelled (continuous).

%We propose that training on composite image maximise the usage of training data as all reference data could be used since cloud are gap-filled

\textit{Predict on single-date images and build consensus predictions.} This paper demonstrated that a model can be trained on a composite image and can generalise to  cloud-free single-date images, which is convenient because most space agencies are now providing surface reflectance data for download. Nonetheless, we highly recommend building consensus by averaging predictions from multiple dates to increase the robustness and confidence of the field extraction process. Our results indicate that averaging the averaging predictions from four images well-spread across the season safeguards against temporal variations of accuracy. Cloud-contaminated images can also be used for inference provided input images with clouds and shadows are properly discarded.  \\

\textit{Optimise thresholds locally.} Fine-tuning thresholds during post-processing  had a significant impact when applying the model to previously-unseen locations. Unlike training data, which are to be collected in blocks, reference data to adjust thresholds should be distributed across the region of interest and cover a range of field types. Optimal threshold values can then be determined locally, \eg with moving windows. \\

These evidence-based principles provide practical guidance for  organising field boundary extraction across vast areas with an automated, data-driven approach and minimum image preprocessing requirements.

\subsection{Perspectives}

% limitation and future plans
In advancing the ability to retrieve individual fields from satellite imagery, our work established that semantic segmentation combined with multitasking is a state-of-the-art approach. It also indicates future direction of research. Foremost among these is testing other image segmentation approaches, such as instance segmentation, where individual labels are assigned to objects of the same class, \eg with Faster Regional Convolutional Neural Network (R-CNN) or  Mask R-CNN~\citep{ren2015faster, he2017mask}.  These models have the potential to produce closed boundaries in a single pass, obviating any post-processing of the semantic segmentation outputs. However, empirical evidence showed that the accuracy of  Mask R-CNN is similar to that of a UNet,  and that combining their predictive performance produce more accurate results than either alone~\citep{vuola2019mask}.\\ % (\textit{7th move: other methods need testing but that's for the future}) 

Our work suggests further topics of investigation to better characterise the model's inner working and data requirements. These topics include 1) to identify the minimum viable training set size, which has direct implications cost of collecting reference data, 2) to evaluate the effect of errors  in the training set (\eg missing fields or approximative field outlines) on  accuracy, and 3) and to untangle the relationship between  accuracy and crop growth processes and crop calendars in the areas being mapped. Finally, more work is needed to evaluate our approach in a wider range of cropping systems such as smallholder farming systems. In that case, very high spatial resolution is likely to be required to resolve small fields. 

%a novel feature of our study\\
%the accuracy values of xxx identified is relatively similar to values reported 

%Given that no other studies have examined..., we cannot make direct comparisons.

%there is (no) reason to believe that...

%

%To discussion: Update of boundaries: \citep{ghaffarian2018improved}, update boundaries using cluster-based snake models. as well as to update them for one season to another~\citep{janssen1993methodology, locherbach1998fusing, torre2000agricultural, butenuth2005network, heipke1999towards, oesterle2004case, matikainen2012challenge}.

\section{Conclusion}

The ability to automatically extract field boundaries from satellite imagery is increasingly needed by many providers of digital agriculture services. In this paper, we formulated this problem as a semantic segmentation task and trained a deep convolutional neural network to solve it. Our model relied on multi-tasking and conditioned inference to predict, for each pixel, the probability of belonging to a field, to a field boundary, as well as to predict the distance to the closest boundary. These predictions were then post-processed to achieve instance segmentation and extract individual fields. By exploiting spectral and contextual information, our neural network demonstrated state-of-the-art performance for field boundary detection and good abilities to generalise across space, time, resolutions, and sensors. \\

Our work provides evidence-based principles for field boundary extraction at scale using deep learning: 1) to train models on monthly cloud-free composites to maximise usage of training data;  2) to predict field boundaries using single-date images because these can replace composites used for prediction with marginal loss of accuracy; 3) to build average predictions from at least four images to cope with the temporal variability of accuracy; 4) to use data-driven procedure to optimally combine model outputs. These principles replace arbitrary parameter selection  with data-driven processes and minimise image preprocessing.

%Once the model is trained, the low image preprocessing requirements of our approach  is expected to facilitate its uptake and enable  field-level analysis across broad geographic areas. 

%With the objective of proposing a data-driven method that has low image preprocessing requirements, we framed the problem of field boundary extraction as a semantic segmentation task, which we tackled with a multi-tasked deep convolutional neural network.  Our neural network generates three segmentation masks (an extent mask, a boundary mask, and a  distance mask) which were subsequently combined to generate individual fields.

\section*{Acknowledgements}
We received funding from ``Grains'', a project of the Digiscape Future Science Platform. We are grateful to  Dr Sophie Bontemps and Annelize Collet for sharing their data and. Finally, we would like to thank Dr Robert Brown, Dr Dave Henry, Dr Joel Dabrowski and the three anonymous reviewers for their insightful comments on earlier versions of the paper. 

\section*{References}
\bibliographystyle{elsarticle-harv}
\bibliography{sample.bib}

\begin{thebibliography}{64}
\expandafter\ifx\csname natexlab\endcsname\relax\def\natexlab#1{#1}\fi
\providecommand{\url}[1]{\texttt{#1}}
\providecommand{\href}[2]{#2}
\providecommand{\path}[1]{#1}
\providecommand{\DOIprefix}{doi:}
\providecommand{\ArXivprefix}{arXiv:}
\providecommand{\URLprefix}{URL: }
\providecommand{\Pubmedprefix}{pmid:}
\providecommand{\doi}[1]{\href{http://dx.doi.org/#1}{\path{#1}}}
\providecommand{\Pubmed}[1]{\href{pmid:#1}{\path{#1}}}
\providecommand{\bibinfo}[2]{#2}
\ifx\xfnm\relax \def\xfnm[#1]{\unskip,\space#1}\fi
%Type = Incollection
\bibitem[{Agravat and Raval(2018)}]{agravat2018deep}
\bibinfo{author}{Agravat, R.R.}, \bibinfo{author}{Raval, M.S.},
  \bibinfo{year}{2018}.
\newblock \bibinfo{title}{Deep learning for automated brain tumor segmentation
  in mri images}, in: \bibinfo{booktitle}{Soft Computing Based Medical Image
  Analysis}. \bibinfo{publisher}{Elsevier}, pp. \bibinfo{pages}{183--201}.
%Type = Article
\bibitem[{Belgiu and Csillik(2018)}]{belgiu2018sentinel}
\bibinfo{author}{Belgiu, M.}, \bibinfo{author}{Csillik, O.},
  \bibinfo{year}{2018}.
\newblock \bibinfo{title}{Sentinel-2 cropland mapping using pixel-based and
  object-based time-weighted dynamic time warping analysis}.
\newblock \bibinfo{journal}{Remote sensing of environment}
  \bibinfo{volume}{204}, \bibinfo{pages}{509--523}.
%Type = Article
\bibitem[{Bergstra and Bengio(2012)}]{bergstra2012random}
\bibinfo{author}{Bergstra, J.}, \bibinfo{author}{Bengio, Y.},
  \bibinfo{year}{2012}.
\newblock \bibinfo{title}{Random search for hyper-parameter optimization}.
\newblock \bibinfo{journal}{Journal of Machine Learning Research}
  \bibinfo{volume}{13}, \bibinfo{pages}{281--305}.
%Type = Article
\bibitem[{Blaes et~al.(2005)Blaes, Vanhalle and Defourny}]{blaes2005efficiency}
\bibinfo{author}{Blaes, X.}, \bibinfo{author}{Vanhalle, L.},
  \bibinfo{author}{Defourny, P.}, \bibinfo{year}{2005}.
\newblock \bibinfo{title}{Efficiency of crop identification based on optical
  and sar image time series}.
\newblock \bibinfo{journal}{Remote sensing of environment}
  \bibinfo{volume}{96}, \bibinfo{pages}{352--365}.
%Type = Article
\bibitem[{Borgefors(1986)}]{Borgefors:1986:DTD:17140.17147}
\bibinfo{author}{Borgefors, G.}, \bibinfo{year}{1986}.
\newblock \bibinfo{title}{Distance transformations in digital images}.
\newblock \bibinfo{journal}{Comput. Vision Graph. Image Process.}
  \bibinfo{volume}{34}, \bibinfo{pages}{344--371}.
\newblock \DOIprefix\doi{10.1016/S0734-189X(86)80047-0}.
%Type = Article
\bibitem[{Boughorbel et~al.(2017)Boughorbel, Jarray and
  El-Anbari}]{boughorbel2017optimal}
\bibinfo{author}{Boughorbel, S.}, \bibinfo{author}{Jarray, F.},
  \bibinfo{author}{El-Anbari, M.}, \bibinfo{year}{2017}.
\newblock \bibinfo{title}{Optimal classifier for imbalanced data using matthews
  correlation coefficient metric}.
\newblock \bibinfo{journal}{PloS one} \bibinfo{volume}{12},
  \bibinfo{pages}{e0177678}.
%Type = Article
\bibitem[{Brodrick et~al.(2019)Brodrick, Davies and
  Asner}]{brodrick2019uncovering}
\bibinfo{author}{Brodrick, P.G.}, \bibinfo{author}{Davies, A.B.},
  \bibinfo{author}{Asner, G.P.}, \bibinfo{year}{2019}.
\newblock \bibinfo{title}{Uncovering ecological patterns with convolutional
  neural networks}.
\newblock \bibinfo{journal}{Trends in ecology \& evolution} .
%Type = Article
\bibitem[{Chai et~al.(2019)Chai, Newsam, Zhang, Qiu and Huang}]{chai2019cloud}
\bibinfo{author}{Chai, D.}, \bibinfo{author}{Newsam, S.},
  \bibinfo{author}{Zhang, H.K.}, \bibinfo{author}{Qiu, Y.},
  \bibinfo{author}{Huang, J.}, \bibinfo{year}{2019}.
\newblock \bibinfo{title}{Cloud and cloud shadow detection in landsat imagery
  based on deep convolutional neural networks}.
\newblock \bibinfo{journal}{Remote sensing of environment}
  \bibinfo{volume}{225}, \bibinfo{pages}{307--316}.
%Type = Article
\bibitem[{Chen et~al.(2015a)Chen, Qiu, Wu and Du}]{chen2015image}
\bibinfo{author}{Chen, B.}, \bibinfo{author}{Qiu, F.}, \bibinfo{author}{Wu,
  B.}, \bibinfo{author}{Du, H.}, \bibinfo{year}{2015}a.
\newblock \bibinfo{title}{Image segmentation based on constrained spectral
  variance difference and edge penalty}.
\newblock \bibinfo{journal}{Remote Sensing} \bibinfo{volume}{7},
  \bibinfo{pages}{5980--6004}.
%Type = Article
\bibitem[{Chen et~al.(2015b)Chen, Chen, Liao, Cao, Chen, Chen, He, Han, Peng,
  Lu et~al.}]{chen2015global}
\bibinfo{author}{Chen, J.}, \bibinfo{author}{Chen, J.}, \bibinfo{author}{Liao,
  A.}, \bibinfo{author}{Cao, X.}, \bibinfo{author}{Chen, L.},
  \bibinfo{author}{Chen, X.}, \bibinfo{author}{He, C.}, \bibinfo{author}{Han,
  G.}, \bibinfo{author}{Peng, S.}, \bibinfo{author}{Lu, M.}, et~al.,
  \bibinfo{year}{2015}b.
\newblock \bibinfo{title}{Global land cover mapping at 30 m resolution: A
  pok-based operational approach}.
\newblock \bibinfo{journal}{ISPRS Journal of Photogrammetry and Remote Sensing}
  \bibinfo{volume}{103}, \bibinfo{pages}{7--27}.
%Type = Article
\bibitem[{Chen et~al.(2018a)Chen, Papandreou, Kokkinos, Murphy and
  Yuille}]{chen2018deeplab}
\bibinfo{author}{Chen, L.C.}, \bibinfo{author}{Papandreou, G.},
  \bibinfo{author}{Kokkinos, I.}, \bibinfo{author}{Murphy, K.},
  \bibinfo{author}{Yuille, A.L.}, \bibinfo{year}{2018}a.
\newblock \bibinfo{title}{Deeplab: Semantic image segmentation with deep
  convolutional nets, atrous convolution, and fully connected crfs}.
\newblock \bibinfo{journal}{IEEE transactions on pattern analysis and machine
  intelligence} \bibinfo{volume}{40}, \bibinfo{pages}{834--848}.
%Type = Article
\bibitem[{Chen et~al.(2017)Chen, Papandreou, Schroff and
  Adam}]{chen2017rethinking}
\bibinfo{author}{Chen, L.C.}, \bibinfo{author}{Papandreou, G.},
  \bibinfo{author}{Schroff, F.}, \bibinfo{author}{Adam, H.},
  \bibinfo{year}{2017}.
\newblock \bibinfo{title}{Rethinking atrous convolution for semantic image
  segmentation}.
\newblock \bibinfo{journal}{arXiv preprint arXiv:1706.05587} .
%Type = Article
\bibitem[{Chen et~al.(2018b)Chen, Fan, Yang, Wang and
  Latif}]{chen2018extraction}
\bibinfo{author}{Chen, Y.}, \bibinfo{author}{Fan, R.}, \bibinfo{author}{Yang,
  X.}, \bibinfo{author}{Wang, J.}, \bibinfo{author}{Latif, A.},
  \bibinfo{year}{2018}b.
\newblock \bibinfo{title}{Extraction of urban water bodies from high-resolution
  remote-sensing imagery using deep learning}.
\newblock \bibinfo{journal}{Water} \bibinfo{volume}{10}, \bibinfo{pages}{585}.
%Type = Article
\bibitem[{Cheng et~al.(2017)Cheng, Wang, Xu, Wang, Xiang and
  Pan}]{cheng2017automatic}
\bibinfo{author}{Cheng, G.}, \bibinfo{author}{Wang, Y.}, \bibinfo{author}{Xu,
  S.}, \bibinfo{author}{Wang, H.}, \bibinfo{author}{Xiang, S.},
  \bibinfo{author}{Pan, C.}, \bibinfo{year}{2017}.
\newblock \bibinfo{title}{Automatic road detection and centerline extraction
  via cascaded end-to-end convolutional neural network}.
\newblock \bibinfo{journal}{IEEE Transactions on Geoscience and Remote Sensing}
  \bibinfo{volume}{55}, \bibinfo{pages}{3322--3337}.
%Type = Article
\bibitem[{Coello(2000)}]{coello2000updated}
\bibinfo{author}{Coello, C.A.}, \bibinfo{year}{2000}.
\newblock \bibinfo{title}{An updated survey of ga-based multiobjective
  optimization techniques}.
\newblock \bibinfo{journal}{ACM Computing Surveys (CSUR)} \bibinfo{volume}{32},
  \bibinfo{pages}{109--143}.
%Type = Misc
\bibitem[{{Crop Estimates Consortium}(2017)}]{SAfieldboundary}
\bibinfo{author}{{Crop Estimates Consortium}}, \bibinfo{year}{2017}.
\newblock \bibinfo{title}{{Field Crop Boundary data layer}}.
%Type = Article
\bibitem[{De~Wit and Clevers(2004)}]{de2004efficiency}
\bibinfo{author}{De~Wit, A.}, \bibinfo{author}{Clevers, J.},
  \bibinfo{year}{2004}.
\newblock \bibinfo{title}{Efficiency and accuracy of per-field classification
  for operational crop mapping}.
\newblock \bibinfo{journal}{International journal of remote sensing}
  \bibinfo{volume}{25}, \bibinfo{pages}{4091--4112}.
%Type = Article
\bibitem[{Defourny et~al.(2019)Defourny, Bontemps, Bellemans, Cara, Dedieu,
  Guzzonato, Hagolle, Inglada, Nicola, Rabaute et~al.}]{defourny2019near}
\bibinfo{author}{Defourny, P.}, \bibinfo{author}{Bontemps, S.},
  \bibinfo{author}{Bellemans, N.}, \bibinfo{author}{Cara, C.},
  \bibinfo{author}{Dedieu, G.}, \bibinfo{author}{Guzzonato, E.},
  \bibinfo{author}{Hagolle, O.}, \bibinfo{author}{Inglada, J.},
  \bibinfo{author}{Nicola, L.}, \bibinfo{author}{Rabaute, T.}, et~al.,
  \bibinfo{year}{2019}.
\newblock \bibinfo{title}{Near real-time agriculture monitoring at national
  scale at parcel resolution: Performance assessment of the sen2-agri automated
  system in various cropping systems around the world}.
\newblock \bibinfo{journal}{Remote sensing of environment}
  \bibinfo{volume}{221}, \bibinfo{pages}{551--568}.
%Type = Article
\bibitem[{Diakogiannis et~al.(2019)Diakogiannis, Waldner, Caccetta and
  Wu}]{diakogiannis2019resunet}
\bibinfo{author}{Diakogiannis, F.I.}, \bibinfo{author}{Waldner, F.},
  \bibinfo{author}{Caccetta, P.}, \bibinfo{author}{Wu, C.},
  \bibinfo{year}{2019}.
\newblock \bibinfo{title}{Resunet-a: a deep learning framework for semantic
  segmentation of remotely sensed data}.
\newblock \bibinfo{journal}{arXiv preprint arXiv:1904.00592} .
%Type = Article
\bibitem[{Egorov et~al.(2019)Egorov, Roy, Zhang, Li, Yan and
  Huang}]{egorov2019landsat}
\bibinfo{author}{Egorov, A.V.}, \bibinfo{author}{Roy, D.P.},
  \bibinfo{author}{Zhang, H.K.}, \bibinfo{author}{Li, Z.},
  \bibinfo{author}{Yan, L.}, \bibinfo{author}{Huang, H.}, \bibinfo{year}{2019}.
\newblock \bibinfo{title}{Landsat 4, 5 and 7 (1982 to 2017) analysis ready data
  (ard) observation coverage over the conterminous united states and
  implications for terrestrial monitoring}.
\newblock \bibinfo{journal}{Remote Sensing} \bibinfo{volume}{11},
  \bibinfo{pages}{447}.
%Type = Article
\bibitem[{Evans et~al.(2002)Evans, Jones, Svalbe and
  Berman}]{evans2002segmenting}
\bibinfo{author}{Evans, C.}, \bibinfo{author}{Jones, R.},
  \bibinfo{author}{Svalbe, I.}, \bibinfo{author}{Berman, M.},
  \bibinfo{year}{2002}.
\newblock \bibinfo{title}{Segmenting multispectral landsat tm images into field
  units}.
\newblock \bibinfo{journal}{IEEE Transactions on Geoscience and Remote Sensing}
  \bibinfo{volume}{40}, \bibinfo{pages}{1054--1064}.
%Type = Article
\bibitem[{Garcia-Pedrero et~al.(2017)Garcia-Pedrero, Gonzalo-Mart{\'\i}n and
  Lillo-Saavedra}]{garcia2017machine}
\bibinfo{author}{Garcia-Pedrero, A.}, \bibinfo{author}{Gonzalo-Mart{\'\i}n,
  C.}, \bibinfo{author}{Lillo-Saavedra, M.}, \bibinfo{year}{2017}.
\newblock \bibinfo{title}{A machine learning approach for agricultural parcel
  delineation through agglomerative segmentation}.
\newblock \bibinfo{journal}{International journal of remote sensing}
  \bibinfo{volume}{38}, \bibinfo{pages}{1809--1819}.
%Type = Book
\bibitem[{Goodfellow et~al.(2016)Goodfellow, Bengio and
  Courville}]{goodfellow2016deep}
\bibinfo{author}{Goodfellow, I.}, \bibinfo{author}{Bengio, Y.},
  \bibinfo{author}{Courville, A.}, \bibinfo{year}{2016}.
\newblock \bibinfo{title}{Deep learning}.
\newblock \bibinfo{publisher}{MIT press}.
%Type = Article
\bibitem[{Graesser and Ramankutty(2017)}]{graesser2017detection}
\bibinfo{author}{Graesser, J.}, \bibinfo{author}{Ramankutty, N.},
  \bibinfo{year}{2017}.
\newblock \bibinfo{title}{Detection of cropland field parcels from landsat
  imagery}.
\newblock \bibinfo{journal}{Remote Sensing of Environment}
  \bibinfo{volume}{201}, \bibinfo{pages}{165--180}.
%Type = Article
\bibitem[{Hagolle et~al.(2010)Hagolle, Huc, Pascual and
  Dedieu}]{hagolle2010multi}
\bibinfo{author}{Hagolle, O.}, \bibinfo{author}{Huc, M.},
  \bibinfo{author}{Pascual, D.V.}, \bibinfo{author}{Dedieu, G.},
  \bibinfo{year}{2010}.
\newblock \bibinfo{title}{A multi-temporal method for cloud detection, applied
  to formosat-2, ven$\mu$s, landsat and sentinel-2 images}.
\newblock \bibinfo{journal}{Remote Sensing of Environment}
  \bibinfo{volume}{114}, \bibinfo{pages}{1747--1755}.
%Type = Inproceedings
\bibitem[{He et~al.(2017)He, Gkioxari, Doll{\'a}r and Girshick}]{he2017mask}
\bibinfo{author}{He, K.}, \bibinfo{author}{Gkioxari, G.},
  \bibinfo{author}{Doll{\'a}r, P.}, \bibinfo{author}{Girshick, R.},
  \bibinfo{year}{2017}.
\newblock \bibinfo{title}{Mask r-cnn}, in: \bibinfo{booktitle}{Proceedings of
  the IEEE international conference on computer vision}, pp.
  \bibinfo{pages}{2961--2969}.
%Type = Inproceedings
\bibitem[{He et~al.(2016a)He, Zhang, Ren and Sun}]{he2016identity}
\bibinfo{author}{He, K.}, \bibinfo{author}{Zhang, X.}, \bibinfo{author}{Ren,
  S.}, \bibinfo{author}{Sun, J.}, \bibinfo{year}{2016}a.
\newblock \bibinfo{title}{Identity mappings in deep residual networks}, in:
  \bibinfo{booktitle}{European conference on computer vision},
  \bibinfo{organization}{Springer}. pp. \bibinfo{pages}{630--645}.
%Type = Article
\bibitem[{He et~al.(2016b)He, Zhang, Ren and Sun}]{DBLP:journals/corr/HeZR016}
\bibinfo{author}{He, K.}, \bibinfo{author}{Zhang, X.}, \bibinfo{author}{Ren,
  S.}, \bibinfo{author}{Sun, J.}, \bibinfo{year}{2016}b.
\newblock \bibinfo{title}{Identity mappings in deep residual networks}.
\newblock \bibinfo{journal}{CoRR} \bibinfo{volume}{abs/1603.05027}.
\newblock \href{http://arxiv.org/abs/1603.05027}{{\tt arXiv:1603.05027}}.
%Type = Article
\bibitem[{Inglada et~al.(2017)Inglada, Vincent, Arias, Tardy, Morin and
  Rodes}]{inglada2017operational}
\bibinfo{author}{Inglada, J.}, \bibinfo{author}{Vincent, A.},
  \bibinfo{author}{Arias, M.}, \bibinfo{author}{Tardy, B.},
  \bibinfo{author}{Morin, D.}, \bibinfo{author}{Rodes, I.},
  \bibinfo{year}{2017}.
\newblock \bibinfo{title}{Operational high resolution land cover map production
  at the country scale using satellite image time series}.
\newblock \bibinfo{journal}{Remote Sensing} \bibinfo{volume}{9},
  \bibinfo{pages}{95}.
%Type = Article
\bibitem[{Ioffe and Szegedy(2015)}]{ioffe2015batch}
\bibinfo{author}{Ioffe, S.}, \bibinfo{author}{Szegedy, C.},
  \bibinfo{year}{2015}.
\newblock \bibinfo{title}{Batch normalization: Accelerating deep network
  training by reducing internal covariate shift}.
\newblock \bibinfo{journal}{arXiv preprint arXiv:1502.03167} .
%Type = Article
\bibitem[{Isikdogan et~al.(2018)Isikdogan, Bovik and
  Passalacqua}]{isikdogan2018learning}
\bibinfo{author}{Isikdogan, F.}, \bibinfo{author}{Bovik, A.},
  \bibinfo{author}{Passalacqua, P.}, \bibinfo{year}{2018}.
\newblock \bibinfo{title}{Learning a river network extractor using an adaptive
  loss function}.
\newblock \bibinfo{journal}{IEEE Geoscience and Remote Sensing Letters}
  \bibinfo{volume}{15}, \bibinfo{pages}{813--817}.
%Type = Article
\bibitem[{Ji et~al.(2018)Ji, Zhang, Xu, Shi and Duan}]{ji20183d}
\bibinfo{author}{Ji, S.}, \bibinfo{author}{Zhang, C.}, \bibinfo{author}{Xu,
  A.}, \bibinfo{author}{Shi, Y.}, \bibinfo{author}{Duan, Y.},
  \bibinfo{year}{2018}.
\newblock \bibinfo{title}{3d convolutional neural networks for crop
  classification with multi-temporal remote sensing images}.
\newblock \bibinfo{journal}{Remote Sensing} \bibinfo{volume}{10},
  \bibinfo{pages}{75}.
%Type = Article
\bibitem[{Kingma and Ba(2014)}]{DBLP:journals/corr/KingmaB14}
\bibinfo{author}{Kingma, D.P.}, \bibinfo{author}{Ba, J.}, \bibinfo{year}{2014}.
\newblock \bibinfo{title}{Adam: {A} method for stochastic optimization}.
\newblock \bibinfo{journal}{CoRR} \bibinfo{volume}{abs/1412.6980}.
\newblock \URLprefix \url{http://arxiv.org/abs/1412.6980},
  \href{http://arxiv.org/abs/1412.6980}{{\tt arXiv:1412.6980}}.
%Type = Article
\bibitem[{Lesiv et~al.(2019)Lesiv, Laso~Bayas, See, Duerauer, Dahlia, Durando,
  Hazarika, Kumar~Sahariah, Vakolyuk, Blyshchyk et~al.}]{lesiv2019estimating}
\bibinfo{author}{Lesiv, M.}, \bibinfo{author}{Laso~Bayas, J.C.},
  \bibinfo{author}{See, L.}, \bibinfo{author}{Duerauer, M.},
  \bibinfo{author}{Dahlia, D.}, \bibinfo{author}{Durando, N.},
  \bibinfo{author}{Hazarika, R.}, \bibinfo{author}{Kumar~Sahariah, P.},
  \bibinfo{author}{Vakolyuk, M.}, \bibinfo{author}{Blyshchyk, V.}, et~al.,
  \bibinfo{year}{2019}.
\newblock \bibinfo{title}{Estimating the global distribution of field size
  using crowdsourcing}.
\newblock \bibinfo{journal}{Global change biology} \bibinfo{volume}{25},
  \bibinfo{pages}{174--186}.
%Type = Misc
\bibitem[{Massey et~al.(2017)Massey, Sankey, Yadav, Congalton, Tilton and
  Thenkabail}]{massey2017NASA}
\bibinfo{author}{Massey, R.}, \bibinfo{author}{Sankey, T.},
  \bibinfo{author}{Yadav, K.}, \bibinfo{author}{Congalton, R.},
  \bibinfo{author}{Tilton, J.}, \bibinfo{author}{Thenkabail, P.},
  \bibinfo{year}{2017}.
\newblock \bibinfo{title}{Nasa making earth system data records for use in
  research environments (measures) global food security-support analysis data
  (gfsad) cropland extent 2010 north america 30 m v001 [data set]}.
\newblock
  \bibinfo{howpublished}{\url{http://dx.doi.org/10.5067/MEaSUREs/GFSAD/GFSAD30NACE.001}}.
\newblock \DOIprefix\doi{10.5067/MEaSUREs/GFSAD/GFSAD30NACE.001}.
%Type = Article
\bibitem[{Matthews(1975)}]{matthews1975comparison}
\bibinfo{author}{Matthews, B.W.}, \bibinfo{year}{1975}.
\newblock \bibinfo{title}{Comparison of the predicted and observed secondary
  structure of t4 phage lysozyme}.
\newblock \bibinfo{journal}{Biochimica et Biophysica Acta (BBA)-Protein
  Structure} \bibinfo{volume}{405}, \bibinfo{pages}{442--451}.
%Type = Article
\bibitem[{Matton et~al.(2015)Matton, Canto, Waldner, Valero, Morin, Inglada,
  Arias, Bontemps, Koetz and Defourny}]{matton2015automated}
\bibinfo{author}{Matton, N.}, \bibinfo{author}{Canto, G.S.},
  \bibinfo{author}{Waldner, F.}, \bibinfo{author}{Valero, S.},
  \bibinfo{author}{Morin, D.}, \bibinfo{author}{Inglada, J.},
  \bibinfo{author}{Arias, M.}, \bibinfo{author}{Bontemps, S.},
  \bibinfo{author}{Koetz, B.}, \bibinfo{author}{Defourny, P.},
  \bibinfo{year}{2015}.
\newblock \bibinfo{title}{An automated method for annual cropland mapping along
  the season for various globally-distributed agrosystems using high spatial
  and temporal resolution time series}.
\newblock \bibinfo{journal}{Remote Sensing} \bibinfo{volume}{7},
  \bibinfo{pages}{13208--13232}.
%Type = Article
\bibitem[{Meyer and Beucher(1990)}]{meyer1990morphological}
\bibinfo{author}{Meyer, F.}, \bibinfo{author}{Beucher, S.},
  \bibinfo{year}{1990}.
\newblock \bibinfo{title}{Morphological segmentation}.
\newblock \bibinfo{journal}{Journal of visual communication and image
  representation} \bibinfo{volume}{1}, \bibinfo{pages}{21--46}.
%Type = Article
\bibitem[{Milletari et~al.(2016)Milletari, Navab and
  Ahmadi}]{DBLP:journals/corr/MilletariNA16}
\bibinfo{author}{Milletari, F.}, \bibinfo{author}{Navab, N.},
  \bibinfo{author}{Ahmadi, S.}, \bibinfo{year}{2016}.
\newblock \bibinfo{title}{V-net: Fully convolutional neural networks for
  volumetric medical image segmentation}.
\newblock \bibinfo{journal}{CoRR} \bibinfo{volume}{abs/1606.04797}.
\newblock \href{http://arxiv.org/abs/1606.04797}{{\tt arXiv:1606.04797}}.
%Type = Article
\bibitem[{Mueller et~al.(2004)Mueller, Segl and Kaufmann}]{mueller2004edge}
\bibinfo{author}{Mueller, M.}, \bibinfo{author}{Segl, K.},
  \bibinfo{author}{Kaufmann, H.}, \bibinfo{year}{2004}.
\newblock \bibinfo{title}{Edge-and region-based segmentation technique for the
  extraction of large, man-made objects in high-resolution satellite imagery}.
\newblock \bibinfo{journal}{Pattern recognition} \bibinfo{volume}{37},
  \bibinfo{pages}{1619--1628}.
%Type = Article
\bibitem[{Novikov et~al.(2017)Novikov, Major, Lenis, Hladuvka, Wimmer and
  B{\"{u}}hler}]{DBLP:journals/corr/NovikovMLHWB17}
\bibinfo{author}{Novikov, A.A.}, \bibinfo{author}{Major, D.},
  \bibinfo{author}{Lenis, D.}, \bibinfo{author}{Hladuvka, J.},
  \bibinfo{author}{Wimmer, M.}, \bibinfo{author}{B{\"{u}}hler, K.},
  \bibinfo{year}{2017}.
\newblock \bibinfo{title}{Fully convolutional architectures for multi-class
  segmentation in chest radiographs}.
\newblock \bibinfo{journal}{CoRR} \bibinfo{volume}{abs/1701.08816}.
\newblock \href{http://arxiv.org/abs/1701.08816}{{\tt arXiv:1701.08816}}.
%Type = Article
\bibitem[{Oliphant et~al.(2019)Oliphant, Thenkabail, Teluguntla, Xiong, Gumma,
  Congalton and Yadav}]{oliphant2019mapping}
\bibinfo{author}{Oliphant, A.J.}, \bibinfo{author}{Thenkabail, P.S.},
  \bibinfo{author}{Teluguntla, P.}, \bibinfo{author}{Xiong, J.},
  \bibinfo{author}{Gumma, M.K.}, \bibinfo{author}{Congalton, R.G.},
  \bibinfo{author}{Yadav, K.}, \bibinfo{year}{2019}.
\newblock \bibinfo{title}{Mapping cropland extent of southeast and northeast
  asia using multi-year time-series landsat 30-m data using a random forest
  classifier on the google earth engine cloud}.
\newblock \bibinfo{journal}{International Journal of Applied Earth Observation
  and Geoinformation} \bibinfo{volume}{81}, \bibinfo{pages}{110--124}.
%Type = Article
\bibitem[{Pan and Yang(2010)}]{Pan:2010:STL:1850483.1850545}
\bibinfo{author}{Pan, S.J.}, \bibinfo{author}{Yang, Q.}, \bibinfo{year}{2010}.
\newblock \bibinfo{title}{A survey on transfer learning}.
\newblock \bibinfo{journal}{IEEE Trans. on Knowl. and Data Eng.}
  \bibinfo{volume}{22}, \bibinfo{pages}{1345--1359}.
\newblock \URLprefix \url{http://dx.doi.org/10.1109/TKDE.2009.191},
  \DOIprefix\doi{10.1109/TKDE.2009.191}.
%Type = Inproceedings
\bibitem[{Penatti et~al.(2015)Penatti, Nogueira and dos Santos}]{7301382}
\bibinfo{author}{Penatti, O.A.}, \bibinfo{author}{Nogueira, K.},
  \bibinfo{author}{dos Santos, J.A.}, \bibinfo{year}{2015}.
\newblock \bibinfo{title}{Do deep features generalize from everyday objects to
  remote sensing and aerial scenes domains?}, in: \bibinfo{booktitle}{2015 IEEE
  Conference on Computer Vision and Pattern Recognition Workshops (CVPRW)}, pp.
  \bibinfo{pages}{44--51}.
\newblock \URLprefix
  \url{doi.ieeecomputersociety.org/10.1109/CVPRW.2015.7301382},
  \DOIprefix\doi{10.1109/CVPRW.2015.7301382}.
%Type = Article
\bibitem[{Persello and Bruzzone(2010)}]{persello2010novel}
\bibinfo{author}{Persello, C.}, \bibinfo{author}{Bruzzone, L.},
  \bibinfo{year}{2010}.
\newblock \bibinfo{title}{A novel protocol for accuracy assessment in
  classification of very high resolution images}.
\newblock \bibinfo{journal}{IEEE Transactions on Geoscience and Remote Sensing}
  \bibinfo{volume}{48}, \bibinfo{pages}{1232--1244}.
%Type = Article
\bibitem[{Persello et~al.(2019)Persello, Tolpekin, Bergado and
  de~By}]{persello2019delineation}
\bibinfo{author}{Persello, C.}, \bibinfo{author}{Tolpekin, V.},
  \bibinfo{author}{Bergado, J.}, \bibinfo{author}{de~By, R.},
  \bibinfo{year}{2019}.
\newblock \bibinfo{title}{Delineation of agricultural fields in smallholder
  farms from satellite images using fully convolutional networks and
  combinatorial grouping}.
\newblock \bibinfo{journal}{Remote Sensing of Environment}
  \bibinfo{volume}{231}, \bibinfo{pages}{111253}.
%Type = Misc
\bibitem[{Phalke et~al.(2017)Phalke, Ozdogan, Thenkabail, Congalton, Yadav,
  Massey, Teluguntla and Smith}]{phalke2017NASA}
\bibinfo{author}{Phalke, A.}, \bibinfo{author}{Ozdogan, M.},
  \bibinfo{author}{Thenkabail, S., P.G.}, \bibinfo{author}{Congalton, R.},
  \bibinfo{author}{Yadav, K.}, \bibinfo{author}{Massey, R.},
  \bibinfo{author}{Teluguntla, P., P.J.}, \bibinfo{author}{Smith, C.},
  \bibinfo{year}{2017}.
\newblock \bibinfo{title}{Nasa making earth system data records for use in
  research environments (measures) global food security-support analysis data
  (gfsad) cropland extent 2015 europe, central asia, russia, middle east 30 m
  v001 [data set]}.
\newblock
  \bibinfo{howpublished}{\url{http://dx.doi.org/10.5067/MEaSUREs/GFSAD/GFSAD30EUCEARUMECE.001}}.
\newblock \DOIprefix\doi{10.5067/MEaSUREs/GFSAD/GFSAD30EUCEARUMECE.001}.
%Type = Inproceedings
\bibitem[{Ren et~al.(2015)Ren, He, Girshick and Sun}]{ren2015faster}
\bibinfo{author}{Ren, S.}, \bibinfo{author}{He, K.}, \bibinfo{author}{Girshick,
  R.}, \bibinfo{author}{Sun, J.}, \bibinfo{year}{2015}.
\newblock \bibinfo{title}{Faster r-cnn: Towards real-time object detection with
  region proposal networks}, in: \bibinfo{booktitle}{Advances in neural
  information processing systems}, pp. \bibinfo{pages}{91--99}.
%Type = Inproceedings
\bibitem[{Ronneberger et~al.(2015)Ronneberger, Fischer and
  Brox}]{ronneberger2015u}
\bibinfo{author}{Ronneberger, O.}, \bibinfo{author}{Fischer, P.},
  \bibinfo{author}{Brox, T.}, \bibinfo{year}{2015}.
\newblock \bibinfo{title}{U-net: Convolutional networks for biomedical image
  segmentation}, in: \bibinfo{booktitle}{International Conference on Medical
  image computing and computer-assisted intervention},
  \bibinfo{organization}{Springer}. pp. \bibinfo{pages}{234--241}.
%Type = Article
\bibitem[{Ruder(2017)}]{ruder2017overview}
\bibinfo{author}{Ruder, S.}, \bibinfo{year}{2017}.
\newblock \bibinfo{title}{An overview of multi-task learning in deep neural
  networks}.
\newblock \bibinfo{journal}{arXiv preprint arXiv:1706.05098} .
%Type = Article
\bibitem[{Rydberg and Borgefors(2001)}]{rydberg2001integrated}
\bibinfo{author}{Rydberg, A.}, \bibinfo{author}{Borgefors, G.},
  \bibinfo{year}{2001}.
\newblock \bibinfo{title}{Integrated method for boundary delineation of
  agricultural fields in multispectral satellite images}.
\newblock \bibinfo{journal}{IEEE Transactions on Geoscience and Remote Sensing}
  \bibinfo{volume}{39}, \bibinfo{pages}{2514--2520}.
%Type = Article
\bibitem[{Salman(2006)}]{salman2006image}
\bibinfo{author}{Salman, N.}, \bibinfo{year}{2006}.
\newblock \bibinfo{title}{Image segmentation based on watershed and edge
  detection techniques.}
\newblock \bibinfo{journal}{Int. Arab J. Inf. Technol.} \bibinfo{volume}{3},
  \bibinfo{pages}{104--110}.
%Type = Article
\bibitem[{Shrivakshan and Chandrasekar(2012)}]{shrivakshan2012comparison}
\bibinfo{author}{Shrivakshan, G.}, \bibinfo{author}{Chandrasekar, C.},
  \bibinfo{year}{2012}.
\newblock \bibinfo{title}{A comparison of various edge detection techniques
  used in image processing}.
\newblock \bibinfo{journal}{International Journal of Computer Science Issues
  (IJCSI)} \bibinfo{volume}{9}, \bibinfo{pages}{269}.
%Type = Article
\bibitem[{Soille and Ansoult(1990)}]{soille1990automated}
\bibinfo{author}{Soille, P.J.}, \bibinfo{author}{Ansoult, M.M.},
  \bibinfo{year}{1990}.
\newblock \bibinfo{title}{Automated basin delineation from digital elevation
  models using mathematical morphology}.
\newblock \bibinfo{journal}{Signal Processing} \bibinfo{volume}{20},
  \bibinfo{pages}{171--182}.
%Type = Article
\bibitem[{Turker and Kok(2013)}]{turker2013field}
\bibinfo{author}{Turker, M.}, \bibinfo{author}{Kok, E.H.},
  \bibinfo{year}{2013}.
\newblock \bibinfo{title}{Field-based sub-boundary extraction from remote
  sensing imagery using perceptual grouping}.
\newblock \bibinfo{journal}{ISPRS journal of photogrammetry and remote sensing}
  \bibinfo{volume}{79}, \bibinfo{pages}{106--121}.
%Type = Article
\bibitem[{Vuola et~al.(2019)Vuola, Akram and Kannala}]{vuola2019mask}
\bibinfo{author}{Vuola, A.O.}, \bibinfo{author}{Akram, S.U.},
  \bibinfo{author}{Kannala, J.}, \bibinfo{year}{2019}.
\newblock \bibinfo{title}{Mask-rcnn and u-net ensembled for nuclei
  segmentation}.
\newblock \bibinfo{journal}{arXiv preprint arXiv:1901.10170} .
%Type = Article
\bibitem[{Waldner et~al.(2018)Waldner, Duveiller and
  Defourny}]{waldner2018local}
\bibinfo{author}{Waldner, F.}, \bibinfo{author}{Duveiller, G.},
  \bibinfo{author}{Defourny, P.}, \bibinfo{year}{2018}.
\newblock \bibinfo{title}{Local adjustments of image spatial resolution to
  optimize large-area mapping in the era of big data}.
\newblock \bibinfo{journal}{International journal of applied earth observation
  and geoinformation} \bibinfo{volume}{73}, \bibinfo{pages}{374--385}.
%Type = Article
\bibitem[{Waldner et~al.(2017)Waldner, Hansen, Potapov, L{\"o}w, Newby,
  Ferreira and Defourny}]{waldner2017national}
\bibinfo{author}{Waldner, F.}, \bibinfo{author}{Hansen, M.C.},
  \bibinfo{author}{Potapov, P.V.}, \bibinfo{author}{L{\"o}w, F.},
  \bibinfo{author}{Newby, T.}, \bibinfo{author}{Ferreira, S.},
  \bibinfo{author}{Defourny, P.}, \bibinfo{year}{2017}.
\newblock \bibinfo{title}{National-scale cropland mapping based on
  spectral-temporal features and outdated land cover information}.
\newblock \bibinfo{journal}{PloS one} \bibinfo{volume}{12},
  \bibinfo{pages}{e0181911}.
%Type = Article
\bibitem[{Watkins and van Niekerk(2019)}]{watkins2019comparison}
\bibinfo{author}{Watkins, B.}, \bibinfo{author}{van Niekerk, A.},
  \bibinfo{year}{2019}.
\newblock \bibinfo{title}{A comparison of object-based image analysis
  approaches for field boundary delineation using multi-temporal sentinel-2
  imagery}.
\newblock \bibinfo{journal}{Computers and Electronics in Agriculture}
  \bibinfo{volume}{158}, \bibinfo{pages}{294--302}.
%Type = Article
\bibitem[{Wilcoxon(1945)}]{wilcoxon1945individual}
\bibinfo{author}{Wilcoxon, F.}, \bibinfo{year}{1945}.
\newblock \bibinfo{title}{Individual comparisons by ranking methods}.
\newblock \bibinfo{journal}{Biometrics Bulletin} \bibinfo{volume}{1},
  \bibinfo{pages}{80--83}.
%Type = Article
\bibitem[{Yan and Roy(2014)}]{yan2014automated}
\bibinfo{author}{Yan, L.}, \bibinfo{author}{Roy, D.}, \bibinfo{year}{2014}.
\newblock \bibinfo{title}{Automated crop field extraction from multi-temporal
  web enabled landsat data}.
\newblock \bibinfo{journal}{Remote Sensing of Environment}
  \bibinfo{volume}{144}, \bibinfo{pages}{42--64}.
%Type = Article
\bibitem[{Zhan et~al.(2005)Zhan, Molenaar, Tempfli and Shi}]{zhan2005quality}
\bibinfo{author}{Zhan, Q.}, \bibinfo{author}{Molenaar, M.},
  \bibinfo{author}{Tempfli, K.}, \bibinfo{author}{Shi, W.},
  \bibinfo{year}{2005}.
\newblock \bibinfo{title}{Quality assessment for geo-spatial objects derived
  from remotely sensed data}.
\newblock \bibinfo{journal}{International Journal of Remote Sensing}
  \bibinfo{volume}{26}, \bibinfo{pages}{2953--2974}.
%Type = Inproceedings
\bibitem[{Zhao et~al.(2017)Zhao, Shi, Qi, Wang and Jia}]{zhao2017pyramid}
\bibinfo{author}{Zhao, H.}, \bibinfo{author}{Shi, J.}, \bibinfo{author}{Qi,
  X.}, \bibinfo{author}{Wang, X.}, \bibinfo{author}{Jia, J.},
  \bibinfo{year}{2017}.
\newblock \bibinfo{title}{Pyramid scene parsing network}, in:
  \bibinfo{booktitle}{Proceedings of the IEEE conference on computer vision and
  pattern recognition}, pp. \bibinfo{pages}{2881--2890}.
%Type = Misc
\bibitem[{Zhong et~al.(2017)Zhong, Giri, Thenkabail, Teluguntla, Congalton,
  Yadav, Oliphant, Xiong, Poehnelt and Smith}]{zhong2017NASA}
\bibinfo{author}{Zhong, Y.}, \bibinfo{author}{Giri, C.},
  \bibinfo{author}{Thenkabail, P.}, \bibinfo{author}{Teluguntla, P.},
  \bibinfo{author}{Congalton, G., R.}, \bibinfo{author}{Yadav, K.},
  \bibinfo{author}{Oliphant, J., A.}, \bibinfo{author}{Xiong, J.},
  \bibinfo{author}{Poehnelt, J.}, \bibinfo{author}{Smith, C.},
  \bibinfo{year}{2017}.
\newblock \bibinfo{title}{Nasa making earth system data records for use in
  research environments (measures) global food security-support analysis data
  (gfsad) cropland extent 2015 south america 30 m v001 [data set]}.
\newblock
  \bibinfo{howpublished}{\url{http://dx.doi.org/10.5067/MEaSUREs/GFSAD/GFSAD30SACE.001}}.
\newblock \DOIprefix\doi{10.5067/MEaSUREs/GFSAD/GFSAD30SACE.001}.

\end{thebibliography}






% \end{thebibliography}

\end{document}